\documentclass[runningheads]{llncs}

\usepackage{eccv}

\usepackage{eccvabbrv}

\usepackage{graphicx}
\usepackage{booktabs}

\usepackage[accsupp]{axessibility}  

\usepackage{hyperref}
\usepackage{enumitem}
\usepackage{booktabs}
\usepackage{arydshln}
\usepackage{caption} 
\usepackage{multirow} 
\usepackage{enumitem}
\usepackage{colortbl} 
\usepackage{bbm}
\usepackage{setspace}                       
\usepackage{lipsum} 
\usepackage{subcaption} 
\usepackage{xcolor}
\usepackage[ruled,linesnumbered]{algorithm2e}

\usepackage{mathtools} 
\usepackage{bm} 
\usepackage{soul} 
\usepackage{nicefrac}
\usepackage{csquotes}
\usepackage{array}
\usepackage{duckuments}
\usepackage{wrapfig} 
\usepackage{orcidlink}

\begin{document}

\title{Dual-End Consistency Model} 
\titlerunning{DE-CM}

\def\spaces{~~~~~~}
\author{
Linwei Dong\textsuperscript{1,2}\spaces{}
Ruoyu Guo\textsuperscript{2}$^{\dagger}$\spaces{}
Ge Bai\textsuperscript{2}\spaces{}
Zehuan Yuan\textsuperscript{2}\spaces{} \\
Yawei Luo\textsuperscript{1}\thanks{Corresponding Authors. ~~\textsuperscript{\dag} Project Leader.}\spaces{}
Changqing Zou\textsuperscript{1,3}\\
}

\authorrunning{Linwei Dong et al.}
\institute{
\textsuperscript{1}Zhejiang University\spaces{}
\textsuperscript{2}Bytedance Inc.\spaces{} 
\textsuperscript{3}Zhejiang Lab \\
}

\maketitle

\begin{abstract}
The slow iterative sampling nature remains a major bottleneck for the practical deployment of diffusion and flow-based generative models. While consistency models (CMs) represent a state-of-the-art distillation-based approach for efficient generation, their large-scale application is still limited by two key issues: training instability and inflexible sampling. Existing methods seek to mitigate these problems through architectural adjustments or regularized objectives, yet overlook the critical reliance on trajectory selection. In this work, we first conduct an analysis on these two limitations: training instability originates from loss divergence induced by unstable self-supervised term, whereas sampling inflexibility arises from error accumulation. Based on these insights and analysis, we propose the {\bf Dual-End Consistency Model (DE-CM)} that selects vital sub-trajectory clusters to achieve stable and effective training. DE-CM decomposes the PF-ODE trajectory and selects three critical sub-trajectories as optimization targets. Specifically, our approach leverages continuous-time CMs objectives to achieve few-step distillation and utilizes flow matching as a boundary regularizer to stabilize the training process. Furthermore, we propose a novel noise-to-noisy (N2N) mapping that can map noise to any point, thereby alleviating the error accumulation in the first step. Extensive experimental results show the effectiveness of our method: it achieves a state-of-the-art FID score of 1.70 in one-step generation on the ImageNet 256×256 dataset, outperforming existing CM-based one-step approaches.
\end{abstract}    
\section{Introduction}
\label{sec:intro}
\begin{figure*}[!t]
  \centering
    \includegraphics[width=\linewidth]{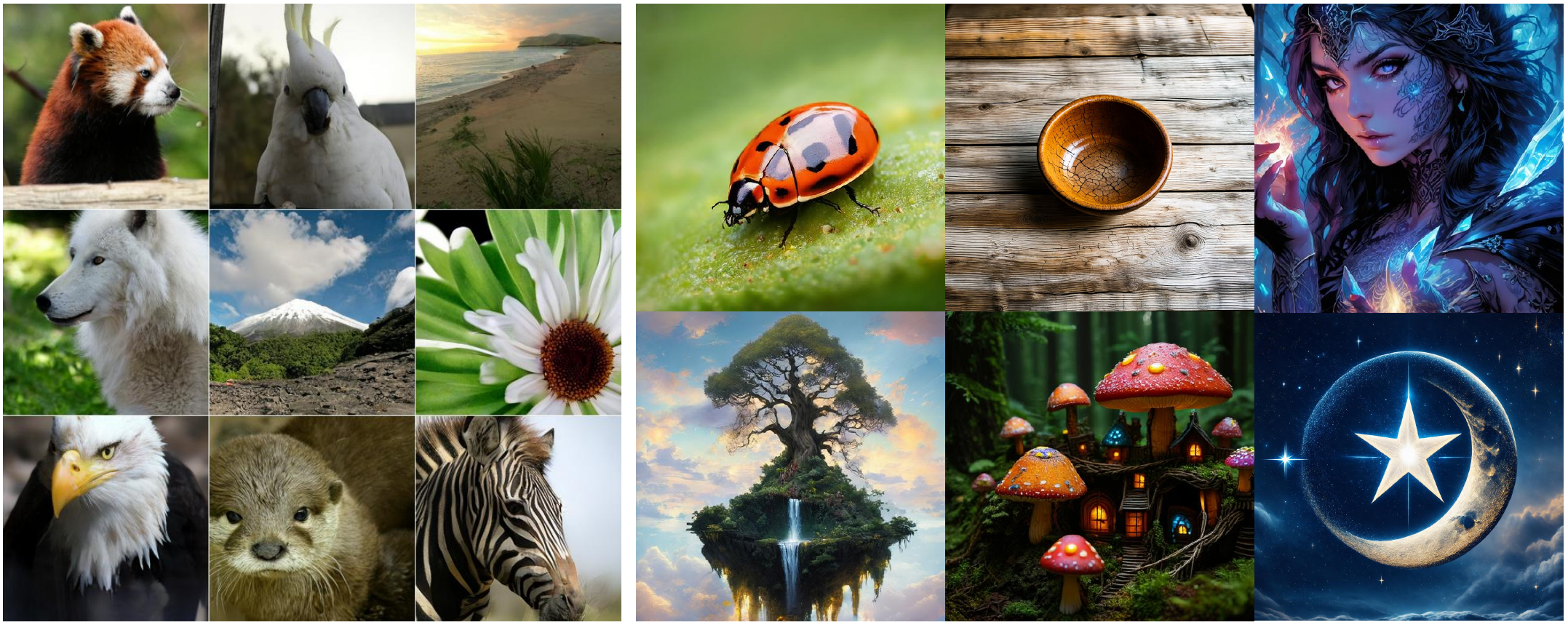}
   \caption{\textbf{Left:} Selected samples from DE-CM trained on ImageNet at 256 × 256 resolution with 2 NFE. \textbf{Right:} 
   Images generated by our text-to-image model across 2 to 50 NFE. The prompts used are provided in the Appendix \ref{ap:text_prompt}.
   }
   \label{fig:teaser}
\end{figure*}
The rapid advancement of diffusion \cite{ho2020denoising, song2019generative, song2020score, sohl2015deep} and flow-based \cite{lipman2022flow, liu2022flow, esser2024scaling} generative models has revolutionized data generation across image \cite{ramesh2022hierarchical, rombach2022high, zhang2023adding}, 3D \cite{liu2023zero, poole2022dreamfusion, wang2023prolificdreamer}, audio \cite{liu2023audioldm, evans2024fast} and video generation \cite{blattmann2023stable, wan2025}, yet their reliance on slow iterative sampling remains a fundamental bottleneck. 
While higher-order solvers \cite{lu2022dpm,lu2025dpm, song2020denoising} offer limited acceleration, distillation techniques \cite{sauer2024adversarial, yin2024improved, yin2024one}, particularly consistency models (CMs) \cite{song2023consistency, kim2023consistency},
have emerged as the most effective path toward few-step generation.
CMs achieve efficient distillation by learning consistency function that directly maps noise to data along predefined trajectories.
However, early discrete-time implementations suffered from discretization errors that potentially compromised output quality \cite{song2023improved, geng2024consistency}, ultimately driving the development of continuous-time variants \cite{lu2024simplifying, sabour2025align}. 

\begin{figure}[!t]
  \centering
  \setlength{\abovecaptionskip}{0.1cm}
    \includegraphics[width=\linewidth]{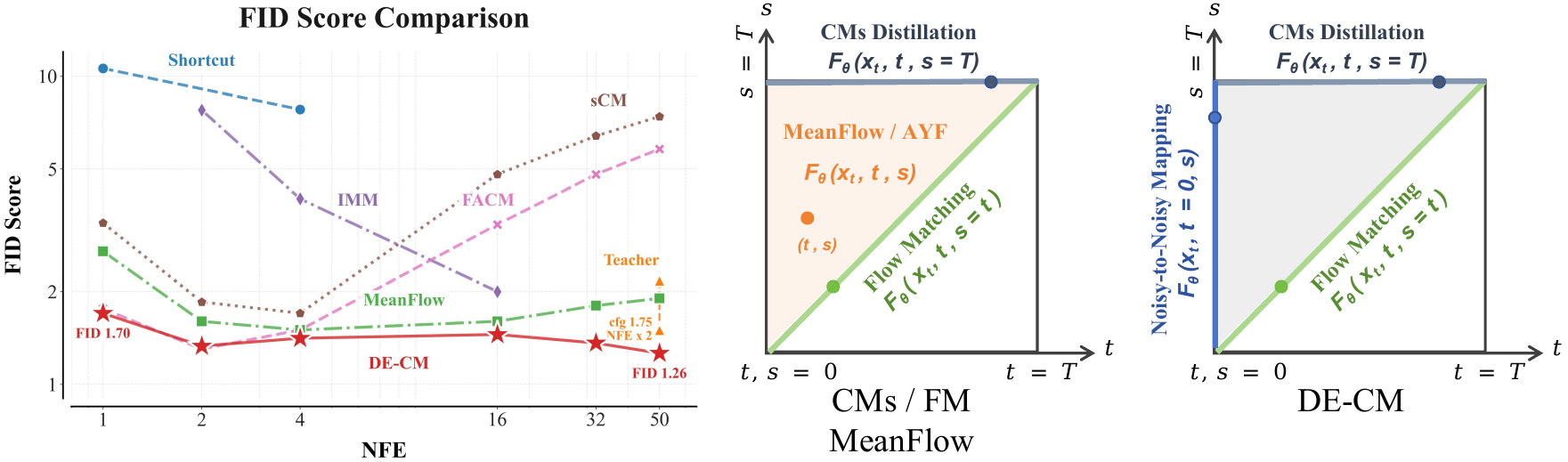}
    \caption{\textbf{Left (a):} Comparison of FID scores
across different
models under various NFE settings, showing the
superior performance of our method in both few-
step and multi-step sampling. \textbf{Right (b):} Comparison of learning objectives.  
     CMs Distillation \cite{song2023consistency} ($s = T$),
    Flow Matching \cite{lipman2022flow} ($s = t$), 
    MeanFlow \cite{geng2025mean} / AYF \cite{sabour2025align} (upper triangle). DE-CM (triangular boundary).}
  \label{fig:square}
\end{figure}

Continuous-time CMs, while circumventing discrete error accumulation, do not address the fundamental limitations (including training instability and inflexible sampling) of CMs, thus impeding their wider deployment.
Recent main efforts on stabilization, efficiency, and flexibility fall into three categories: 
1) diffusion modeling. 
Methods like sCM \cite{lu2024simplifying} that employ TrigFlow to simplify the formulation of diffusion models, which unify the flow ODE and CMs.
MeanFlow \cite{geng2025mean}, AYF \cite{sabour2025align} and IMM \cite{zhou2025inductive} introduce the flow maps \cite{boffi2024flow} to map noisy inputs directly to any point along the PF-ODE trajectory.
2) network architecture. 
Approaches \cite{lu2024simplifying} tailor the architecture design to achieve more stable and efficient training.
3) training objective with regularization. 
Approaches like IMM, MeanFlow, SplitMeanFlow  \cite{guo2025splitmeanflow}, sCoT \cite{wu2025traflow} and FACM \cite{peng2025flow} reformulate the learning objective with additional constraints or flow modeling. 
While these efforts facilitate continuous-time CMs training, they remain fundamentally limited by treating suboptimal sub-trajectories as their optimization objectives.
Methods like MeanFlow and AYF couple the optimization throughout the whole trajectories by learning mappings between any two points, as shown in \cref{fig:square} (b). 
These entangled learning objectives are particularly problematic given that CMs impose consistency as a network behavior property without an explicit boundary condition from the ground-truth flow characteristics, leading to slower convergence or even training collapse.
Furthermore, approaches like sCMs treat vanilla CMs trajectory alone, potentially increasing instability due to the absence of the critical path.
By explicitly decoupling the instantaneous velocity objective along the PF-ODE trajectory, FACM achieves improved training stability.
This indicates that the core challenge lies in selecting the appropriate candidate trajectories to achieve both training stability and strong performance.
 
In this work, we begin with an analysis of two inherent limitations in vanilla CMs: unstable training and inflexible sampling. 
Our findings indicate that training instability arises from loss divergence, which is triggered by unstable self-supervised objectives and the catastrophic forgetting of original instantaneous velocity features. 
Furthermore, we attribute inflexible sampling to error accumulation caused by the the CMs sampler \cite{kim2023consistency}.
Based on these analysis, we propose a novel method, {\bf Dual-End Consistency Model}, to achieve efficient, stable, and flexible model distillation under critical sub-trajectory clusters selection.
We employ flow maps to incorporate entire transport space and identify the significant sub-trajectory clusters for optimization.
Taking the intermediate time $t$ as the dividing point, we focus on the following three trajectory segments:
1) Consistency Trajectory. 
2) Instantaneous Trajectory.
3) Noise-to-Noisy Trajectory.
The comparative illustration of our optimization trajectories is provided in \cref{fig:square} (b).
Consistency trajectories enable the model to perform inference over few-step via consistency property, while instantaneous trajectories serve as a boundary regularization to stabilize the training.
By incorporating noise-to-noisy trajectories, we formulate a training objective that enables our model to alleviate the error accumulation imposed by the CMs sampler.
DE-CM excels in few-step sampling on C2I and T2I tasks while maintaining performance across multi-steps, as show in \cref{fig:teaser}.
On ImageNet 256×256 \cite{deng2009imagenet}, DE-CM achieves 1.70 FID using 1-NFE generation, as shown in \cref{fig:square} (a). 

The key contributions are summarized as follows:
\begin{itemize}
    \item We analyze the potential constraints imposed on CMs and elucidate the underlying causes of instability and inflexible sampling.
    \item  We propose DE-CM, a novel method for stable and effective training through the selection of vital trajectories.
    \item We derive a new objective noise-to-noisy mapping to learn a mapping from pure noise to arbitrary points.
    \item Our experiments on C2I and T2I demonstrate the effectiveness of DE-CM, achieving state-of-the-art performance on the ImagNet dataset with 250 training epoch.
\end{itemize}

\section{Preliminaries}
\label{sec:pre}
\paragraph{\bf Diffusion Models and Flow Matching.}
Given a dataset with underlying distribution $p_{data}$  and standard deviation  $\sigma$, diffusion models learn to reverse a noising process that transforms a clean sample $x_1 \sim p_{data}$ into $x_t = \alpha_t x_1 + \sigma_t z$, where $z \sim \mathcal{N}(0, I)$, and the noise level increases as the timestep 
$t$ decreases from $T$ to $0$.
EDM \cite{karras2022elucidating} defines $\alpha_t = 1, \sigma_t = 1 - t$, and optimizes:
\begin{equation}
\mathbb{E}_{x_1, z, t} \left[ w(t) \| f_\theta^{\text{DM}}(x_t, t) - x_1 \|_2^2 \right],
\end{equation}
where $w(t)$ is the weighting and $f_\theta(x_t, t)$ is defined as:
\begin{equation}
    f_\theta(x_t, t) = c_{\text{skip}}(t)x_t + c_{\text{out}}(t)F_\theta(c_{\text{in}}(t)x_t, c_{\text{noise}}(t)),
     \label{eq:edm-f}
\end{equation}
Here, $F_\theta$ is a neural network, and these terms $c(t)$ are scaling coefficients that stabilize training. 

Flow matching (FM) uses differentiable schedules $ \alpha_t, \sigma_t $ (e.g, $ \alpha_t = t, \sigma_t = 1 - t $) and minimizes:
\begin{equation}
    \mathbb{E}_{x_0, z, t} \left[ w(t) \| F_\theta(x_t, t) - (\dot \alpha_t x_0 + \dot \sigma_t z) \|_2^2 \right],
    \label{eq:fm-target}
\end{equation}
where $\dot \alpha_t$ and $\dot \sigma_t$ denote the derivatives with respect to $t$.
Sampling solves the probability flow ODE (PF-ODE):
\begin{equation}
    \frac{dx_t}{dt} = F_\theta(x_t, t),
    \label{eq:fm-ode-sample}
\end{equation}
from $ x_0 \sim \mathcal{N}(0, I) $  at $ t=0 $ to $ t=1 $.

\paragraph{\bf Consistency Models.}
Consistency Models (CMs) are designed to map any point ${x}_t$ along a PF-ODE trajectory directly to its endpoint $x_1$ in a single step. 
This behavior is governed by a self-consistency property: 
for any pair of points ${x}_t$ and $x_{t^\prime}$ on the same trajectory, the model must satisfy $f_\theta({x}_t, t) = f_\theta({x}_{t^\prime}, t^\prime)$. 
This implies $f_\theta(x_t, t) = x_1$ for all $t \in [0, 1]$ with the boundary condition $f_\theta({x}_1, 1) = {x}_1$.

Discrete-time CMs follow the \cref{eq:edm-f} with $c_{skip}(1) = 1, c_{out}(1) = 0$ and learn the consistency between two adjacent timesteps.
The training objective is defined as follows:
\begin{equation}
    \mathbb{E}_{x_1 \sim p_{data}} \left[ w(t) \cdot d \left( f_\theta(x_t, t), f_{\theta^-}(x_{t + \Delta t}, t + \Delta t) \right) \right],
    \label{eq:dis-cm}
\end{equation}
where $w(t)$ is the time-aware weighting,  $\Delta t > 0$ is a minimal step, $\theta^-$ denotes $stopgrad(\theta)$ and $d(\cdot,\cdot)$ is a distance metric.
Discrete-time CMs are sensitive to the discretization step size $\Delta t$, typically requiring carefully designed schedules to ensure convergence. 
Furthermore, the sample $x_{t+\Delta t}$ is commonly generated from $x_t$ by numerically solving the PF-OFE process that introduces additional discretization error, potentially leading to suboptimal sample quality. 

Continuous-time CMs avoid the accumulation of discrete errors by taking the limit $\Delta t \rightarrow 0$.
When employing $d(x,y) = \| x - y \|_2^2 $, the gradient of \cref{eq:dis-cm}  converges to
\begin{equation}
    \nabla_{\theta}\mathbb{E}_{x_{t},t} \left[ -w(t) f_{\theta}^{\top}(x_{t},t) \frac{d f_{\theta^-}(x_{t},t)}{dt} \right],
    \label{eq:con-cm}
\end{equation}
where $\frac{d f_{\theta^-}(\mathbf{x}_t, t)}{d t} = \nabla_{\mathbf{x}_t} f_{\theta^-}(\mathbf{x}_t, t) \frac{d\mathbf{x}_t}{d t} + \partial_t f_{\theta^-}(\mathbf{x}_t, t)$
is the tangent of $f_{\theta^-}$ at $(\mathbf{x}_t, t)$ along the trajectory of the PF-ODE $\frac{d\mathbf{x}_t}{d t}$ and $w(t)$ is the weighting function. 
This modeling approach eliminates the need for a predefined sampling interval $\Delta t$, thus avoiding discretization errors.

\section{Limitations of Consistency Models}
\subsection{Training instability} 
\label{subsec:training-instableity}
\begin{figure}[!t]
  \centering
    \includegraphics[width=\linewidth]{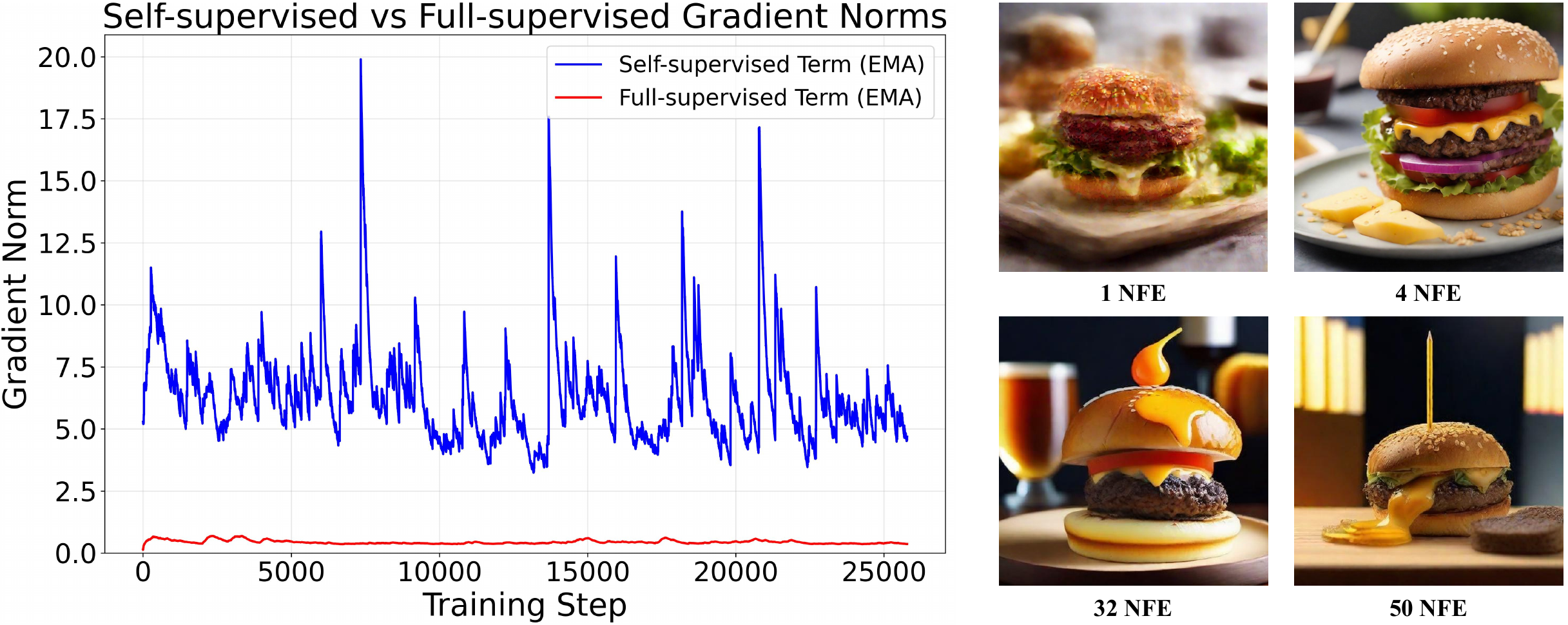}
  \caption{\textbf{Left (a)}: Gradient stability comparison between supervised and self-supervised loss functions. 
  Self-supervised loss exhibits more unstable gradients.
  \textbf{Right (b)}: Visualization results of different NFE sampling using the CM sampler.
  }
  \label{fig:sup-loss}
\end{figure}
Similar to discrete versions, continuous-time CMs face the challenge with training instability \cite{geng2024consistency,song2023improved}.
To better understand the causes of instability, we analyze its ultimate training objective.
Following the naive discrete-time CMs based on OT-FM, we set $c_{skip}(t) = 1$ and $c_{out}(t) = 1 - t$.
Thus the training objective of \cref{eq:con-cm} can be derived as follows:
\begin{equation}
    \mathbb{E}_{x_{t},t} \left [ \| F_{\theta}(x_{t},t)  - F_{\theta^-}(x_{t},t)  +   w^\prime(t) \cdot g_{\theta^-} \|_2^2 \right ] ,
    \label{eq:loss:con-cm-loss}
\end{equation}
where $g_{\theta^-} = F_{\theta^-}(x_t,t) - v_{\phi} - (1 - t)\frac{d F_{\theta^-}(x_t,t)}{dt}$, $v_{\phi}$ denotes the desired trajectory tangent $\frac{dx}{dt}$  and it can be directly derived from data ($v_{\phi} = \dot \alpha x_1 + \dot \sigma x_0$) or predicted by a teacher model ($v_{\phi}= v_\text{tea}(x_t,t)$). $\epsilon$ is equal to $\Delta t$, $w^\prime(t) =  \frac{1}{2}w(t) \epsilon (1 - t)$ and $w(t)$ is a custom weighting function with respect to $t$ and commonly set to $\frac{1}{\epsilon}$ to accelerate convergence.
To clearly delineate the training objectives, we further decompose \cref{eq:loss:con-cm-loss}:
\begin{equation}
    \mathbb{E}_{x_{t},t} \left [ \| F_{\theta}(x_{t},t)  - v_\phi \|_2^2  + \| F_{\theta}(x_{t},t)  - \tilde v   \|_2^2  \right ],
    \label{eq:loss:two-loss}
\end{equation}
where $ \tilde v = F_{\theta^-}(x_{t},t) + (1 - t)\frac{dF_{\theta^-}}{dt}$.
We train the two losses independently, with the gradients generated during training shown in \cref{fig:sup-loss} (a).
Compared to full-supervised term, self-supervised loss is more prone to instability and volatility.
This reveals that:
1) A divergence exists between the fully supervised and self-supervised components when $F_\theta$ has not converged.
2) The emphasis on consistency in $F$ results in the overshadowing of the pretraind instantaneous velocity concept.
These suggest inherent limitations in the optimization trajectory predefined by the vanilla CMs.

\subsection{Inflexible Sampling Mechanism}
\label{subsec:infelxible}
The challenges of inflexible sampling have been thoroughly analyzed in Consistency Trajectory Models (CTM) \cite{kim2023consistency}.
Let $t \in (0,T)$ and $\gamma \in [0,1]$. We denote $p_{\bm{\theta}}$ as the density derived from a sampling trajectory following the transition sequence $T \rightarrow \sqrt{1-\gamma^2}t \rightarrow t \rightarrow 0$, initiated from the prior distribution $p_T$.
Then, we can obtain that
\begin{equation}
    D_{TV}(p_{\text{data}}, p_{\bm{\theta}}) = \mathcal{O}\big(\sqrt{T-\sqrt{1-\gamma^2}t + t}\big), 
\end{equation}
where $D_{TV}(p, q) := \frac{1}{2}\int |p(\mathbf{x}) - q(\mathbf{x})| d\mathbf{x}$.
When generalized to $N$ steps, the sampling behavior is critically governed by the interpolation parameter $\gamma$. For $\gamma=1$, the sampler iteratively performs long jumps from each $t_n$ to $0$, which leads to significant time overlaps between jumps and causes the approximation error to aggregate to $\mathcal{O}(\sqrt{T + t_1 + \cdots + t_N})$. In contrast, setting $\gamma=0$ eliminates these temporal overlaps and prevents error accumulation, resulting in a stable error bound of only $\mathcal{O}(\sqrt{T})$.
Consequently, by leveraging the $\gamma=0$ regime (e.g. Euler Solver), diffusion model can effectively addresses the performance degradation associated with high NFE in distillation models while simultaneously bypassing the discretization errors inherent in score-based models.
However, to satisfy the consistency function and boundary conditions, the sampling mechanism of CMs is based on $\gamma=1$.
The absence of intermediate state transitions imposes the following limitations \cite{kim2023consistency, song2023consistency} on naive CM sampling, explaining its suboptimal performance:
1) Inflexible Sampling Steps.
The model is constrained to a fixed step count, limiting adaptability.
2) Artifact Accumulation. 
Multi-step sampling leads to an accumulation of errors, which often manifests as visual artifacts.
3) Sensitivity to Initial Noise. 
The generation quality is highly sensitive to the choice of initial noise, leading to poor robustness and high output variance.
As shown in \cref{fig:sup-loss} (b), due to the accumulation of errors, the generated quality first increases and then decreases as the NFE gradually increases.
\section{Dual-End Consistency Model}

\begin{figure*}[!t]
  \centering
    \includegraphics[width=\linewidth]{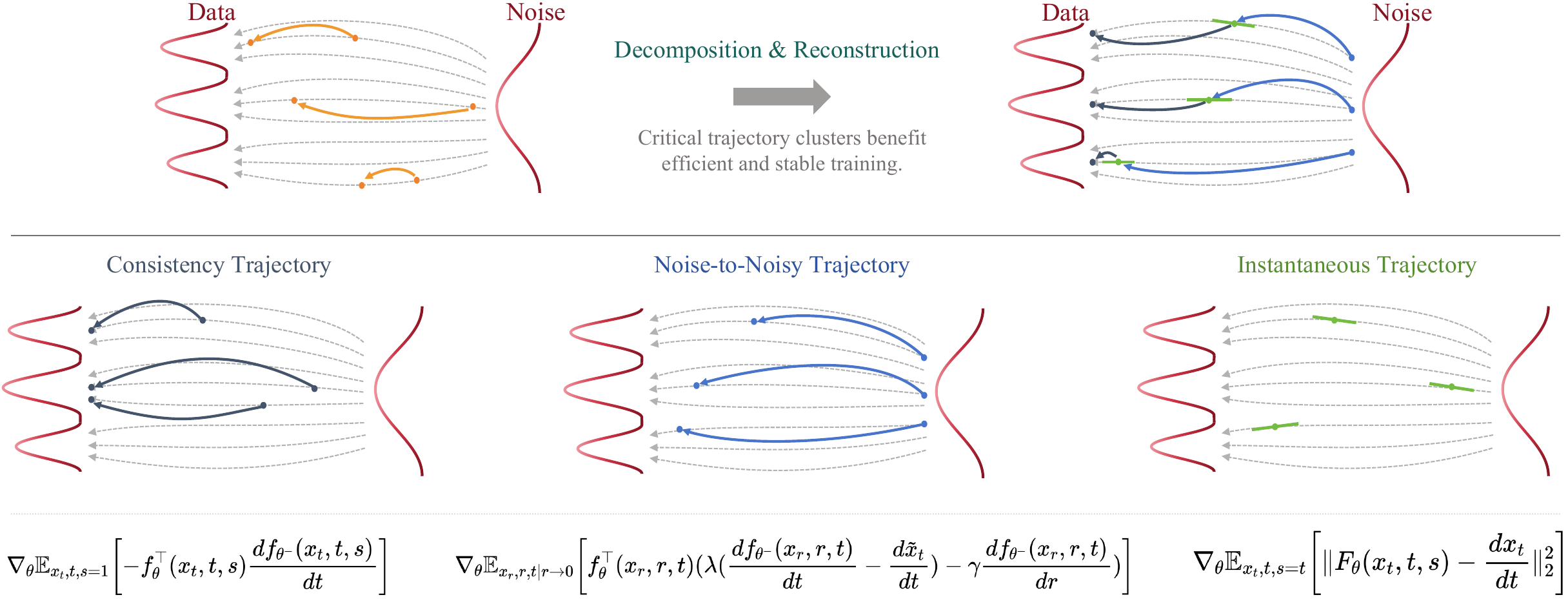}
   \caption{
   We select significant trajectories from the whole $\{(t,s)|t<s \}$ space and treat these selected trajectories as our optimization targets.
   Specifically, we employ the continuous-time consistency distillation trajectory to optimize the mapping from arbitrary time points to data, thereby achieving few-step distillation. 
   We leverage the proposed noise-to-noisy (N2N) mapping objective to eliminate the constraint of predicting only $x_1$ points. 
   We utilize flow matching loss to overcome the instability in training. 
   We find that incorporating these three trajectory clusters is enough to enable stable and efficient optimization for few-step distillation.
   }
   \label{fig:pipeline}
\end{figure*}
MeanFlow \cite{geng2025mean} and AYF \cite{sabour2025align} indicate that the aforementioned limitation of CMs lies in the exclusive dependence on the time condition $t$, which is inherently insufficient to characterize the entire PF-ODE trajectory.
Consequently, they augment the temporal specification with an auxiliary endpoint embedding $s$, enabling the estimation of the mean velocity along the trajectory.
The trajectory consistency function is formulated as follows:
\begin{equation}
    f_{\theta}(x_t,t,s) = x_t + (s-t)F_{\theta}(x_t,t,s)
    \label{eq:f_func}
\end{equation}
where $F_\theta$ is a neural network. 
\cref{eq:f_func} encompasses the entire trajectories when given $t,s \in [0,1]$  and $ t < s$, thus satisfying both CMs and FM parameterization.
This formulation will optimize any sub-trajectory defined by $t$ and $s$ and have access to ODE solver (e.g. Euler Sampler).
However, unlike the consistency function that exhibits explicit boundary conditions $f_\theta({x}_1, 1) = {x}_1$, the boundary conditions of \cref{eq:f_func} is not strictly constrained as $t,s$ are both variables.
This will introduces greater instability risks during training.
Moreover, optimizing the entire sub-trajectory space also leads to inefficient training
as the redundant trajectory optimization degenerates convergence.
The computational complexity, $O(n^2)$, became a primary bottleneck, leading to slow optimization.
Fortunately, we find that optimizing only specific sub-trajectory clusters is sufficient to ensure stable training and high-quality samples.
We focus on three key trajectory clusters (\cref{fig:pipeline}): consistency trajectories, instantaneous trajectories and noise-to-noisy trajectories.

\subsection{Few-Step Models with Consistency Trajectory}
Given $t \in [0,1]$ and $ s = 1 $, \cref{eq:f_func} can be equated to the consistency function of standard CMs. 
By analogy with the derivation of \cref{eq:loss:con-cm-loss}, we can further simplify:
\begin{equation}
    \mathcal{L}_\text{cm} =  \mathbb{E}_{x_{t},t,s|s=1} \left [  \| F_{\theta}(x_{t},t, s) - v_\text{cm}^\text{tar} \|_2^2 \right ],
    \label{eq:loss:simple-con-cm-loss}
\end{equation}
where $v_\text{cm}^\text{tar} = w^\prime(t)\bar v + (1 - w^\prime(t))F_{\theta^-}(x_{t},t, 1)$ and $\bar v = v_\phi + (1 - t) \frac{dF_{\theta^-}(x_t,t,1)}{dt}$.
It can be seen that the final loss is a weighted combination of the model prediction and another velocity $\bar v$.
Moreover, $\bar v$ is defined as the instantaneous velocity $v_\phi(x_t,t)$ corrected by the average acceleration $\frac{dF_{\theta^-}(x_t,t,1)}{dt}$, thus representing the average velocity \cite{geng2025mean}.
Given that $\frac{dF_{\theta^-}(x_t,t,1)}{dt}$ serves as the training objective but is not available in advance, the standard approach is to utilize Jacobian-Vector product (JVP) to dynamically compute it based on current neural network states. 
The total derivative, expanded via the chain rule, is:
\begin{equation}
    \frac{d F_{\theta^-}}{dt} = [\nabla_{x_t} F_{\theta^-}, \frac{\partial {F}_{\theta^-}}{\partial t}, 
    \frac{\partial {F}_{\theta^-}}{\partial s}] 
    \cdot  [v_\phi, 1, 0]^{\top},
    \label{eq:derivate}
\end{equation}
where $[v_\phi, 1, 0]$ is drawn from $[\frac{dx_t}{dt}, \frac{dt}{dt}, \frac{ds}{dt} ]$, and $\frac{dx_t}{dt} = v_\phi$ represents the direction given by the pretrained diffusion or flow model along the PF-ODE.
The consistency properties enable the model to learn the mapping from noisy input $x_t$ to data $x_1$. 
This is analogous to reflow that construct paired data, alleviating the difficulty of learning few-step models.

\subsection{Flexible Models with Noise-to-Noisy Mapping}
By optimizing the time interval $[0, t]$, 
we enable the model to map from pure noise $x_0$ to any intermediate noisy state $x_t$, rather than returning only to the data $x_1$.
This helps mitigate the error accumulation caused by learning only long jumps, by exposing the model to noise inputs similar to those encountered during inference.
To preserve the meaning of the random variable $t$, we reset the left endpoint in \cref{eq:f_func} by $r \rightarrow 0$, while the right remains $t$.
Following \cref{eq:dis-cm}, for timestep $t$, under the $L_2$ norm, we have:
\begin{equation}
    \mathbb{E}_{x_r,r,t|r \rightarrow 0} \left[ w(r,t) \| f_\theta(x_r, r, t), \mathcal{S_\psi}(f_{\theta^-}(x_r, r, t^\prime))  \|_2^2  \right],
    \label{eq:dis-cm-right}
\end{equation}
where $\mathcal{S_\psi}$ refers to \textit{Euler Solver} on the PF-ODE starting from $x_{t^\prime}$  to $x_{t}$ ($t > t^\prime$).
Inspired by \cite{sabour2025align, wu2025traflow}, its gradient can be derived to converge to (See \ref{ap:n2n})
\begin{equation}
    \nabla_{\theta}\mathbb{E}_{x_{r},r,t|r \rightarrow 0} \left[ w(r,t)^\prime f_{\theta}^{\top}(x_{r},r,t) (\frac{d f_{\theta^-}(x_{r},r,t)}{dt} - v_\psi) \right]
    \label{eq:riht-t}
\end{equation}
where $v_\psi = v_\text{tea}(f_{\theta^-}(x_r,r,t),t) $ denotes the desired trajectory tangent $\frac{d \tilde x}{dt}$.
\cref{eq:riht-t} reveals the gradient convergence objective of the self-supervised loss \cref{eq:dis-cm-right} with respect to the right endpoint $t$.
Considering the left endpoint $r$, we can regard it as a consistency function based on boundary $f_{\theta}(x_t,t,t) = x_t$.
Then, its gradient objective can be derived from \cref{eq:con-cm}.
The two optimization objectives can be expressed in weighted form, and their gradients can be proven to converge to (See \ref{ap:n2n})
\begin{equation}
\begin{aligned}
    & \nabla_{\theta}\mathbb{E}_{x_{r},r,t|r \rightarrow 0} 
    \left[ w(r,t)^\prime f_{\theta}^{\top}(x_{r},r,t) 
    (\lambda \frac{d f_{\theta^-}(x_{r},r,t)}{dt} - \right. \\
    & \left.
    \gamma \frac{df_{\theta^-}(x_r,r,t)}{dr} - \lambda \cdot v_\text{tea}(f_{\theta^-}(x_r,r,t),t)) 
    \right],
\end{aligned}
\label{eq:gradient:II}
\end{equation}
where $\lambda, \gamma$ are weighting factors.
The final optimization loss $\mathcal{L}_\text{n2n}$ can be expressed as:
\begin{equation}
   \mathbb{E}_{x_{r},r,t|r \rightarrow 0} \left [ \| F_{\theta}(x_{t},t, s) -  F_{\theta^-}(x_{t},t, s) + w^\prime(r,t)g \|_2^2 \right ],
    \label{eq:loss:II loss}
\end{equation}
where 
$g = (\lambda + \gamma)F_{\theta^-}  - (\gamma v_\phi + \lambda v_\psi ) - (t-r) \dot F_{\theta^-} $ and $ \dot{F}_{\theta -} = [\nabla_{x_t} F_{\theta^-}, \frac{\partial {F}_{\theta^-}}{\partial t}, \frac{\partial {F}_{\theta^-}}{\partial s}] \cdot  [\lambda v_\phi, \lambda, -\gamma]^{\top}.$
\cref{eq:loss:II loss} utilizes the velocities on both sides of the endpoint $r$ and $t$, along with the correction term from the neural network.

\subsection{Stable Models with Instantaneous Velocity}
\begin{figure}[!t]
  \centering
  \begin{subfigure}{0.48\linewidth}
    \includegraphics[width=\linewidth]{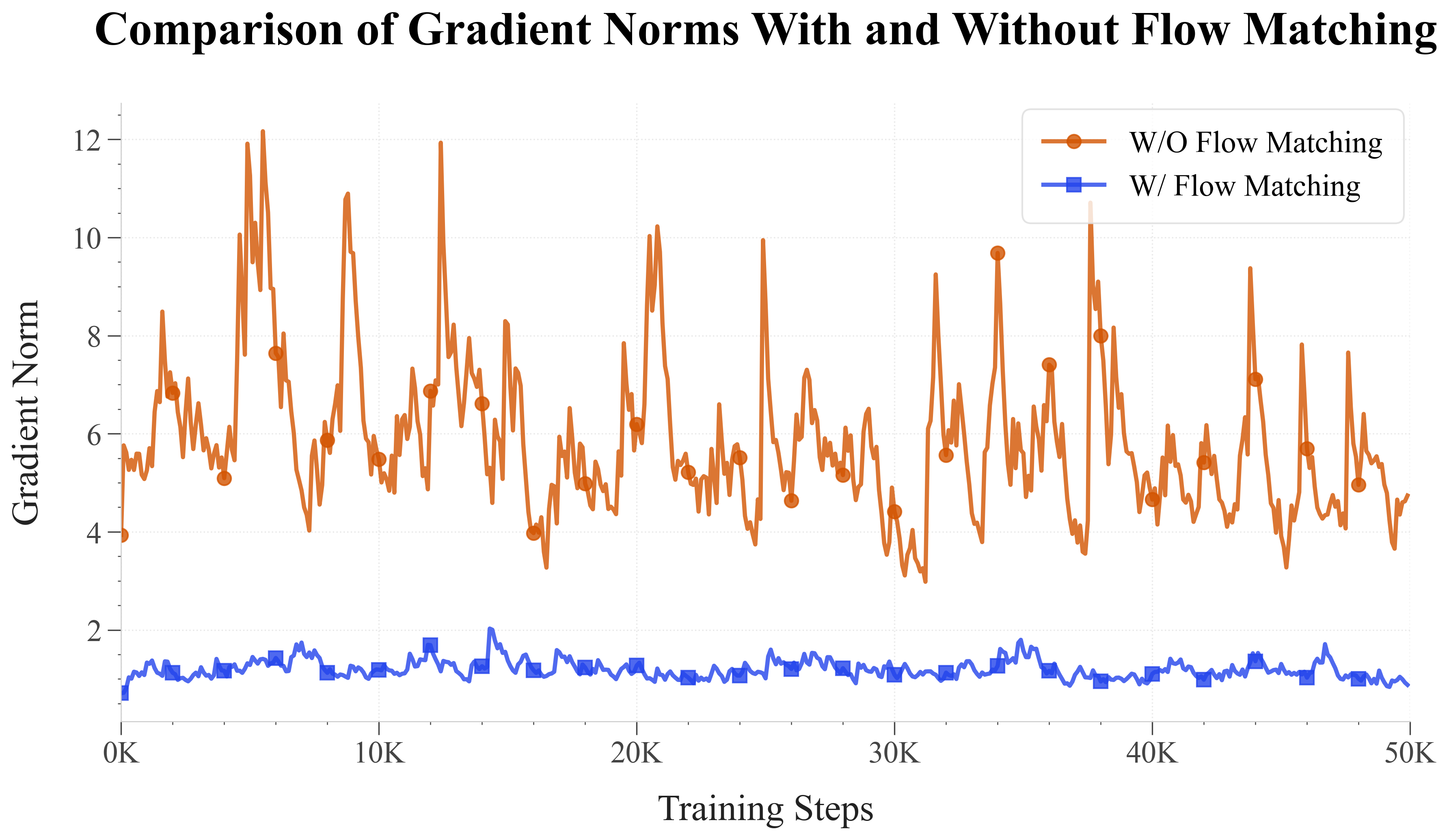}
  \end{subfigure}
  \hfill
    \begin{subfigure}{0.48\linewidth }
    \includegraphics[width=\linewidth]{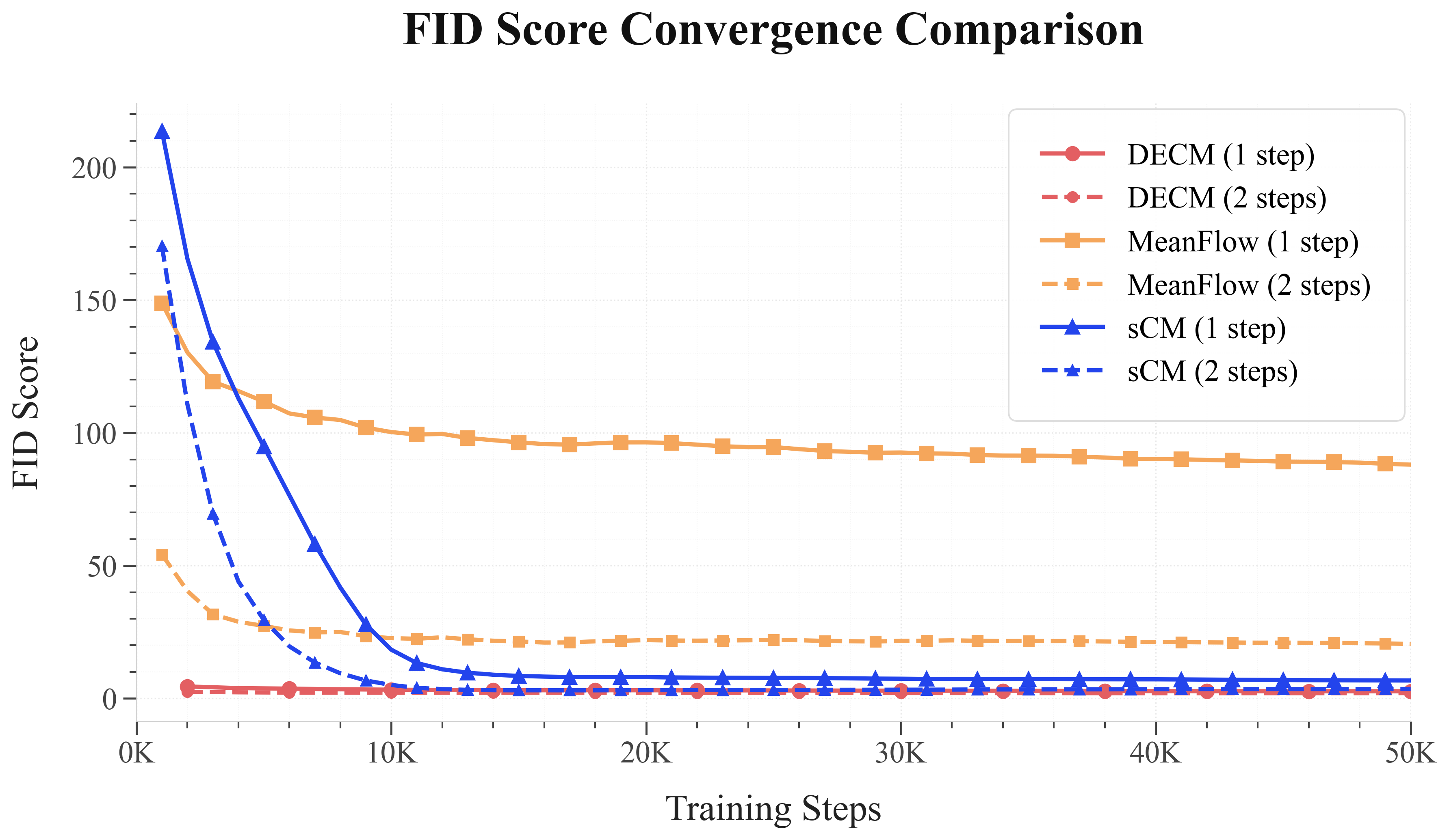}
  \end{subfigure}
   \caption{  \textbf{Left (a):} Comparison of gradient norm curves with and without flow matching boundary constraints.
   \textbf{Right (b):} Training Efficiency Comparison of Different Methods on 8-GPU. Our method achieves a more efficient convergence rate.
   }
   \label{fig:grad_compare}
\end{figure}

Directly using \cref{eq:loss:simple-con-cm-loss} or \cref{eq:loss:II loss} for optimization does not stabilize training.
We find this is due to the absence of necessary boundary conditions.
Following \cref{eq:loss:two-loss}, We can restate the training objective  when given $t$ and $s$:
\begin{equation}
    \mathbb{E}_{x_{t},t,s} \left [ \| F_{\theta}(x_{t},t,s)  - v_\phi \|_2^2  + \| F_{\theta}(x_{t},t,s)  - \tilde v   \|_2^2  \right ],
    \label{eq:loss:two-loss_ts}
\end{equation}
where $\tilde v = F_{\theta^-}(x_{t},t,s) + (s - t)\frac{dF_{\theta^-}}{dt}$.
The dynamic calculation of $\frac{dF_{\theta^-}}{dt}$ in \cref{eq:derivate} renders the self-supervised term susceptible to instability, which raises the risk that online models could diverge from a single erroneous update.
The large gradient phenomena are shown in \cref{fig:grad_compare} (a).

However, when $s=t$, we observe that the original gradient update paradigm is driven solely by ground truth as the second term equals zero. 
This term also represents the left boundary condition.
The single loss term makes the optimization objective deterministic, eliminating slack on the left-hand side.
Similarly, for targets with loose right-hand constraints like \cref{eq:loss:II loss}, a right-hand regularization can be incorporated.
This approach corresponds to selecting all instantaneous trajectory clusters and it makes training more stable (w/ flow in \cref{fig:grad_compare} (a)).
The specific loss $\mathcal{L}_\text{fm}$ is formulated as:
\begin{equation}
    \mathbb{E}_{x_{t},t,s|s=t} \left[ \| F_\theta(x_t,t,s) - v_\phi \|_2^2 + L_{cos}(F_\theta(x_t,t,s), v_\phi)\right],
    \label{eq:loss:fm}
\end{equation}
Beyond its role as a robust regularizer, this loss function is designed to safeguard the pre-trained model's capabilities, thus supporting multi-step ODE sampling.

The overall training process is shown in \cref{alg:distillation}.

\begin{algorithm}[!t]
    \caption{\label{alg:distillation}\textbf{DE-CM Training Procedure}}
    \KwIn{Pretrained diffusion model $F_\text{tea}$, online model $F_\theta$, 
    paired dataset $\mathcal{D}=\{z_\text{ref}, y_\text{ref}\}$, hyperparameters $\lambda, \gamma$, weighting function $w(r,t)$. 
    }
    \KwOut{Trained model $F_\theta$, ema model $F_\text{ema}$.}
    
    \tcp{Initialize model}
    $F_\theta \leftarrow \text{copyWeights}(F_\text{tea}),$
    $F_\text{ema} \leftarrow \text{copyWeights}(F_\theta)$

    \While{train}{
        \tcp{Prepare Data}
        Sample batch $z \sim \mathcal{N}(0, \mathbf{I})$ and $(z_\text{ref}, y_\text{ref}) \sim \mathcal{D}$

        Sample timestep $t\sim\mathcal{U}(0,1), r \rightarrow 0$
        
        $z_t \leftarrow (1 - t)z + tz_\text{ref}$,
        $v_t^\text{cond} \leftarrow  F_\text{tea}(z_t,t ; y_\text{ref}), v_t^\text{uncond} \leftarrow F_\text{tea}(z_t,t ; \phi)$

        $v_t \leftarrow v_t^\text{uncond} + w_{cfg} ( v_t^\text{cond} - v_t^\text{uncond})$

        \text{~}

        \tcp{\textbf{\textcolor{blue}{Continuous-time CD}}}
        $\mathcal{L}_\text{cm} \leftarrow \text{ContinuousCDLoss}(F_\theta(z_t,t,1;y_\text{ref}), v_\text{cm}^\text{tar}) $ 
        \hfill \tcp{Eq~\ref{eq:loss:simple-con-cm-loss}}
        
        \text{~}
        
        \tcp{\textbf{\textcolor{blue}{Flow Matching}}}
        
        $\mathcal{L}_\text{fm} \leftarrow \text{FlowMatchingLoss}(F_\theta(z_t,t,t;y_\text{ref}), v_t)$
        \hfill \tcp{Eq~\ref{eq:loss:fm}}
        
        \text{~}

        \tcp{\textbf{\textcolor{blue}{Noise-to-Noisy Mapping} }}

        $z_r \leftarrow (1 - r)z + rz_\text{ref}$
        
        $v_r^\text{cond} \leftarrow  F_\text{tea}(z_r,r ; y_\text{ref}), v_r^\text{uncond} \leftarrow F_\text{tea}(z_r,r ; \phi)$

        $v_r \leftarrow v_r^\text{uncond} + w_{cfg}  ( v_r^\text{cond} - v_r^\text{uncond})$

        $F_\text{n2n}, \nabla_t F_\text{n2n} \leftarrow  \text{JVP}(F_\theta,(x_r, r, t; y_\text{ref}),(\lambda v_r,\lambda,-\gamma))$

        $g_\text{n2n} \leftarrow (\lambda + \gamma) \cdot \text{sg}(F_\text{n2n}) - \lambda(v_r + (t - r) \nabla_t F_\text{n2n}) - \gamma v_t$

        $v_\text{n2n}^\text{tar} \leftarrow \text{sg}(F_\text{n2n}) - w(r,t) \cdot g_\text{n2n} $

        $\mathcal{L}_\text{n2n} \leftarrow \text{MSELoss}(F_\text{n2n}, v_\text{n2n}^\text{tar}) $ 
        \hfill \tcp{Eq~\ref{eq:loss:II loss}}

        $F_\theta, F_\text{ema} \leftarrow \text{update}(F_\theta, \mathcal{L}_\text{n2n} + \mathcal{L}_\text{cm} + \mathcal{L}_\text{fm})$
        
    }
\end{algorithm}

\section{Experiments}
\subsection{Setup}
For class-to-image (C2I) generation, we use ImageNet-256 $ \times $ 256 datasets \cite{deng2009imagenet}. 
We implement our models in the latent space of a pre-trained VAE tokenizer \cite{yao2025reconstruction}, which produces a latent dimensionality of 32$\times$16$\times$16 .
Building upon the pre-trained LightningDiT \cite{yao2025reconstruction}, we introduce an additional input $s$ to $F_\theta$ to condition the model on the right timestep.
We evaluate using Fréchet Inception Distance (FID) \cite{heusel2017gans} on 50K generated images.
And we train the C2I model with learning rate 1e-4 and AdamW \cite{loshchilov2017decoupled} optimizer.

For text-to-image (T2I) generation, we use 100K data from text-to-image-2M datasets \cite{jackyhatetexttoimag2M}. 
We utilize the default SD3-VAE tokenizer.
We leverage a reward model \cite{xu2023imagereward} to assess generation quality, incorporating CLIP score \cite{radford2021learning}, BLIP score \cite{li2022blip}, and ImageReward \cite{xu2023imagereward}.
We adopt LoRA \cite{hu2022lora} with a rank of 64, a learning rate of 5e-4, and the AdamW optimizer.

\subsection{Qualitative and Quantitative Comparison}
\paragraph{\textbf{Class-to-Image Generation.}}
\begin{figure}[!t]
  \centering
    \includegraphics[width=\linewidth]{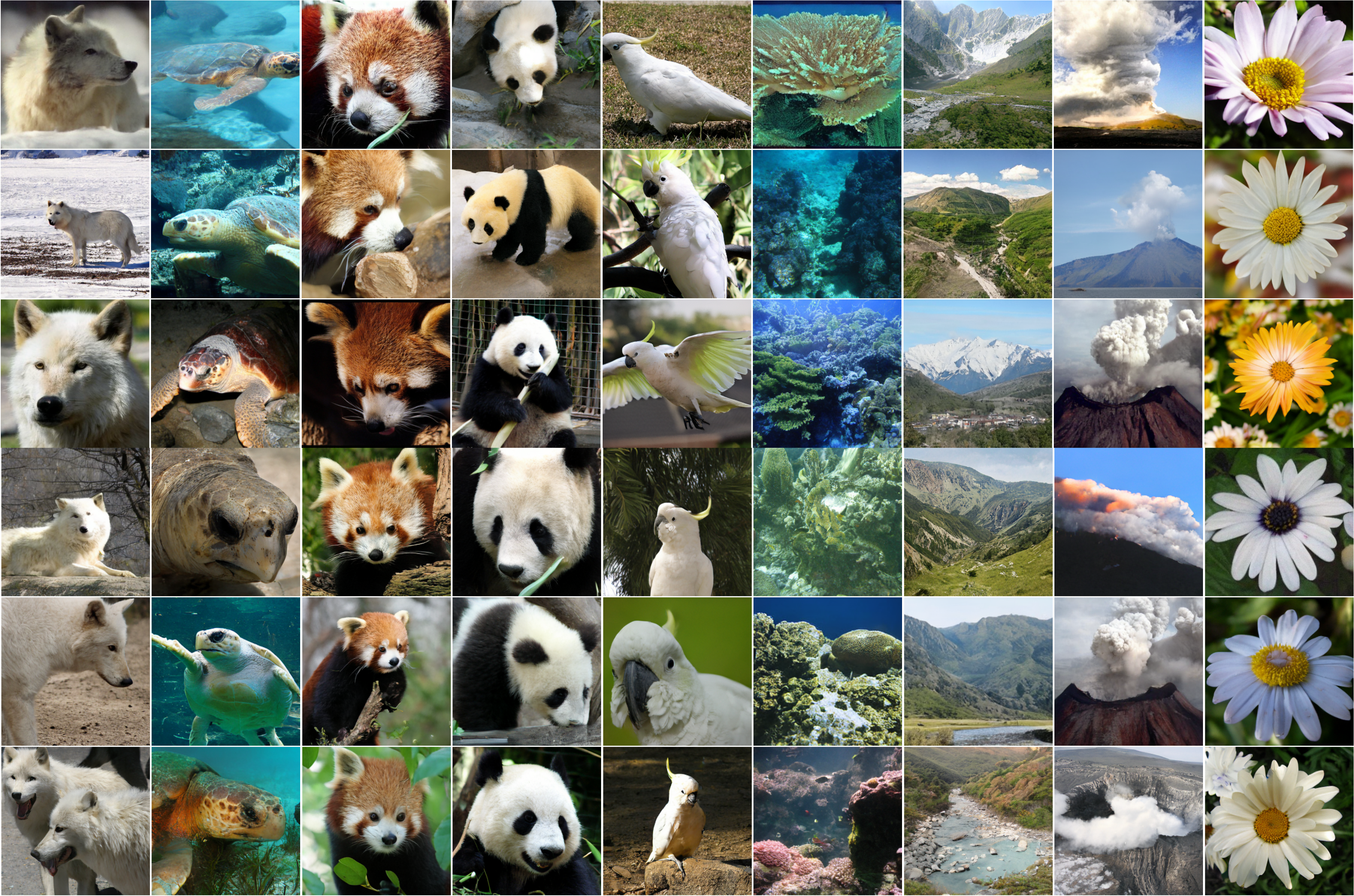}
   \caption{Selected samples from DE-CM trained on ImageNet at 256 × 256 resolution with 1 NFE.
   }
   \label{f8ig:imagent}
\end{figure}

In \cref{tab:img256} we show ImageNet 256 $ \times $ 256 results, reporting FID scores along with the NFE (Number of Function Evaluations). 
Our model establishes a new state-of-the-art at 1 NFE among comparable-scale models and even surpasses the vast majority of multi-step models with just 2 NFE, demonstrating robust and superior performance.
Furthermore, our model maintains stable performance at large NFE using a single set of parameters.
\cref{f8ig:imagent} presents the visual results generated by our model using 1 NFE sampling.

\begin{figure*}[!t]
  \centering
    \includegraphics[width=\linewidth]{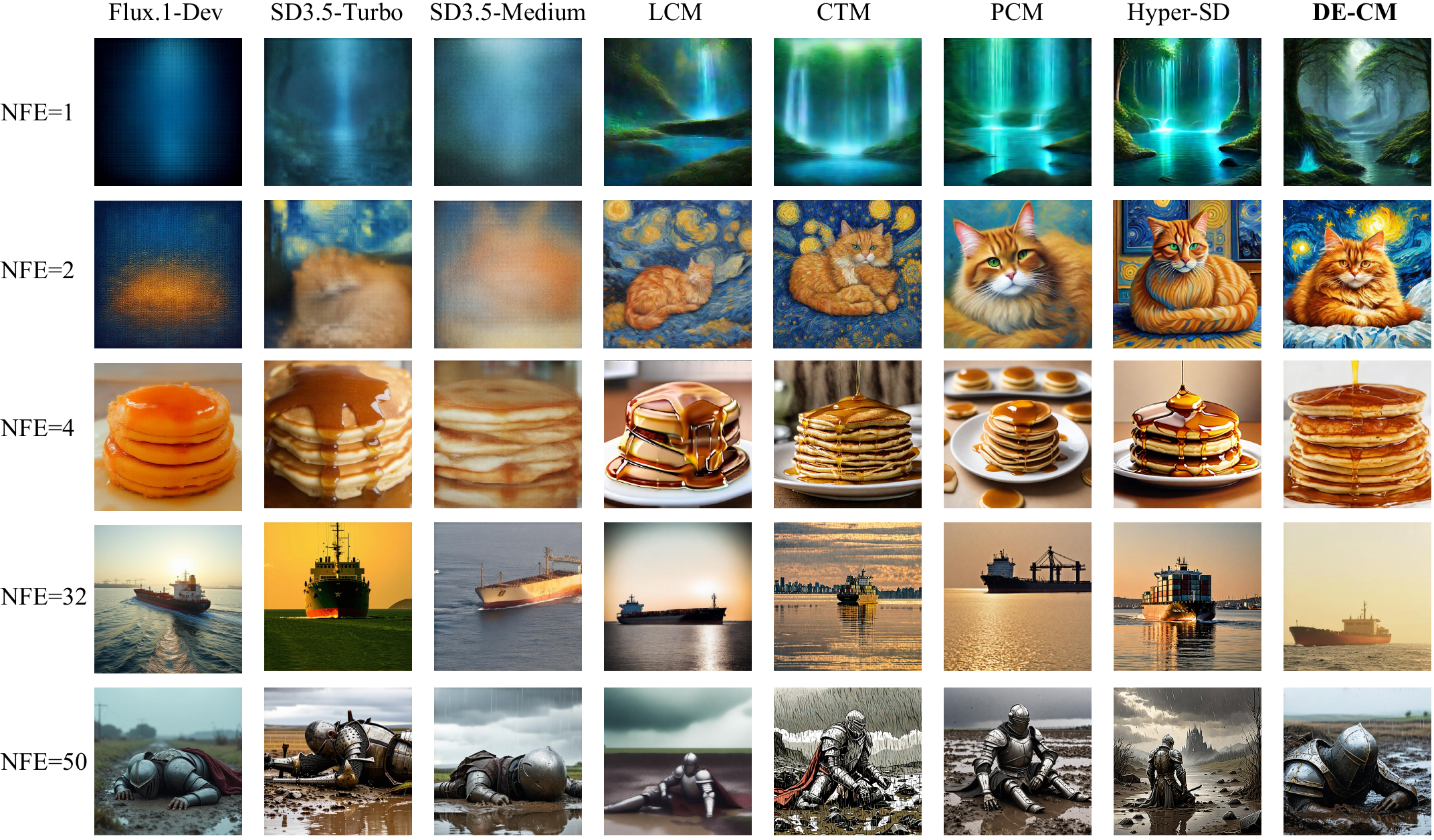}
   \caption{Qualitative visualization reveals DE-CM superiority in quality and text alignment across all NFE. 
   Existing competing models exhibit significant step-quality trade-offs, failing at low NFE (SD3.5-Medium, FLUX.1-Dev) or producing oversaturated samples at high NFE (LCM, CTM, PCM and Hyper-SD). DE-CM maintains robust performance throughout the efficiency-quality spectrum.
   }
   \label{fig:t2i}
\end{figure*}

\paragraph{\textbf{Text-to-Image Generation.}}
\begin{table*}[!t]
    \centering
    \begin{minipage}{0.49\linewidth}
        \centering
        \captionof{table}{Class-conditional generation on ImageNet-256×256. `$\times$ 2' indicates that CFG
doubles the NFE per step. $\dagger$ indicates distillation reproduction (250 epoch) based on our configuration.}
        \label{tab:img256}
        \resizebox{1.0\linewidth}{!}{
            \begin{tabular}{@{}lcccc@{}}
            \toprule
            \textbf{Method} & \textbf{NFE ($\downarrow$)}  & \textbf{\#Params ($\downarrow$)}   & \textbf{Time ($\downarrow$)} & \textbf{FID ($\downarrow$)}\\ 
            \midrule
            \textbf{GAN} & & & & \\ 
            GigaGAN \cite{kang2023scaling} & 1 & 569M &  - & 3.45 \\ 
            StyleGAN-XL \cite{sauer2022stylegan} & 1  & 166M & 0.3 & 2.30 \\ 
            \midrule
            \textbf{Masked \& AR} & & & & \\ 
            VQGAN \cite{esser2021taming} & 256 & 227M & 19 & 3.04 \\
            MaskGIT \cite{chang2022maskgit}  & 8  & 227M & 0.5 & 6.18 \\
            VAR-d30 \cite{tian2024visual} & 10  & 2B & 1 & 1.92 \\
            \midrule
            \textbf{Diffusion \& Flow} & & & & \\ 
            ADM \cite{dhariwal2021diffusion} & 250 & 280M & 16 & 10.94 \\ 
            DiT-XL/2 ($w$=1.5) \cite{peebles2023scalable} & 250 $\times$ 2  & 675M & $>$3.75 & 2.27 \\ 
            SiT-XL/2 ($w = 1.5$)  \cite{ma2024sit} & 250 $\times$ 2 & 675M & $>$  2.89 & 2.15 \\ 
            LightningDiT-XL/1 ($w = 6.7$)  & 250 $\times$ 2  & 675M & $>$  2.82  & 1.35 \\ 
            \midrule
            \textbf{Consistency Model} & & & & \\ 
        MeanFlow \cite{geng2025mean}  & 1  & 675M & 0.011 & 3.43 \\ 
        & 2 &  & 0.024 & 2.20 \\ 
        MeanFlow-$\dagger$ \cite{geng2025mean}  & 1  & 675M & 0.041 & 2.79 \\ 
        & 2 &  & 0.048 & 1.74 \\ 
        IMM \cite{zhou2025inductive}  & 1 $\times$ 2  & 675M & 0.057 & 7.77 \\ 
        & 2 $\times$ 2  &  & 0.068 & 3.99 \\ 
        FACM-$\dagger$ \cite{peng2025flow}  & 1  & 675M & 0.011 & 1.76 \\ 
        iCT \cite{song2023improved}  & 1  & 675M & - & 34.24 \\ 
        & 2  &  & - & 20.30 \\ 
        sCMs-$\dagger$ \cite{lu2024simplifying}  & 1  & 675M & 0.011 & 3.34 \\ 
        & 2 &  & 0.024 & 1.94 \\ 
        Shortcut \cite{frans2024one}  & 1  & 675M  & 0.048 & 10.60 \\ 
        \rowcolor[RGB]{240,230,245}
        \textbf{DE-CM (Ours)}  & 1 & 675M & 0.011 & \textbf{1.70} \\ 
        \rowcolor[RGB]{240,230,245}
         & 2  &  & 0.024 & \textbf{1.33}\\ 
        \rowcolor[RGB]{240,230,245}
         & 50  &  & 0.58 & \textbf{1.26} \\ 
        \bottomrule
        \end{tabular}
        }
    \end{minipage}
    \hfill
    \begin{minipage}{0.49\linewidth}
        \centering
        \captionof{table}{Quantity comparison on pre-trained and CM distillation-based text-to-image model. `$\times$ 2' indicates that CFG
doubles the NFE per step. The best results of each metric are highlighted in \textbf{bold}.}
        \label{tab:t2i}
        \resizebox{1.0\linewidth}{!}{
    \begin{tabular}{@{}lcccc@{}}
\toprule
\textbf{Method} & \textbf{NFE ($\downarrow$)} & \textbf{CLIP ($\uparrow$)} & \textbf{BLIP ($\uparrow$)} & \textbf{ImageRewards ($\uparrow$)} \\
\midrule
\textbf{Multi-step Models} & & & & \\
SD3.5-Medium \cite{esser2024scaling} & 4 & 0.2395 & 0.4457 & -1.6017 \\
& 50 $\times$ 2 & {0.2998} & 0.5451 & 0.8750 \\
SD3.5-Large \cite{esser2024scaling} & 4 & 0.2385 & 0.4221 & -1.582 \\
& 50 $\times$ 2 & \textbf{0.3024} & {0.5465} & 0.8349 \\
Flux.1-Dev \cite{flux.1-dev2024} & 4 & 0.2543 & 0.4523 & -1.2984 \\
& 50 $\times$ 2 & 0.2941 & 0.5384 & {0.9634} \\
\rowcolor[RGB]{240,230,245}
\textbf{DE-CM (Ours)} & 50 & 0.2993 & \textbf{0.5490} & \textbf{0.9712}  \\
\midrule
\textbf{Few-step Models} & & & & \\
\midrule
SD3.5-Turbo \cite{evans2024fast} & 1 & 0.2579 & 0.4974 & -0.8519 \\
& 4 & 0.2953 & {0.5418} & 0.6844 \\
& 8 & 0.2935 & {0.5334} & 0.6513 \\
LCM \cite{luo2023latent} & 1 & 0.2508 & 0.4560 & -1.0622 \\
& 4 & 0.2948 & 0.5306 & 0.4910 \\
& 8 & 0.2993 & 0.5318 & 0.5871 \\
CTM \cite{kim2023consistency} & 1 & 0.2624 & 0.4764 & -0.7639 \\
& 4 & 0.2986 & 0.5284 & 0.5703 \\
& 8 & 0.2993 & 0.5278 & 0.6762 \\
Hyper-SD \cite{ren2024hyper} & 1 & 0.2939 & {0.5161} & {0.5704} \\
& 4 & 0.2949 & 0.5307 & {0.7931} \\
& 8 & \textbf{0.3029} & 0.5293 & {0.8447} \\
PCM \cite{wang2024phased} & 1 & 0.2726 & 0.4764 & -0.6584 \\
& 4 & {0.2981} & 0.5265 & 0.6025 \\
& 8 & 0.2996 & 0.5286 & 0.6195 \\
\midrule
\textbf{DE-CM (Ours)} & 1 & \textbf{0.2996} & \textbf{0.5398} & \textbf{0.5758} \\
\rowcolor[RGB]{240,230,245}
& 4 & \textbf{0.2999} & \textbf{0.5474} & \textbf{0.8117} \\
\rowcolor[RGB]{240,230,245}
& 8 & {0.3000} & \textbf{0.5479} & \textbf{0.8671} \\
\bottomrule
\end{tabular}
        }
    \end{minipage}
\end{table*}
\cref{fig:t2i} provides a visualization of our method against competing models across various inference budgets (1, 2, 4, 32 and 50 NFE).
While pre-trained models like Flux \cite{flux.1-dev2024} and SD3.5 \cite{esser2024scaling} are inherently incapable of few-step generation, existing CMs like LCM \cite{luo2023latent} and CTM \cite{kim2023consistency} suffer from blurry results at low NFE and oversaturation at large NFE.
Methods including PCM \cite{wang2024phased} and Hyper-SD \cite{ren2024hyper}, which require dedicated training for each NFE, face challenges with oversaturation and poor instruction alignment (e.g. fail to render Van Gogh style at 2 NFE).
DE-CM demonstrates compelling performance, achieving appealing results in few-step inference and realistic quality at large NFE.
\cref{fig:t2i-2} further demonstrates the visualization results of our 1 NFE. 
Compared to other methods, our approach exhibits finer textures and enhanced realism.
\cref{tab:t2i} presents a quantitative evaluation using feedback scores.
Our model achieves strong performance across various metrics, notably BLIP scores, and exhibits superior sample quality universally across all NFE settings.

\paragraph{\textbf{Efficiency Comparison.}}
\begin{figure}[!t]
  \centering
  \setlength{\abovecaptionskip}{0.1cm}
  \begin{subfigure}{0.32\linewidth}
    \includegraphics[width=\linewidth]{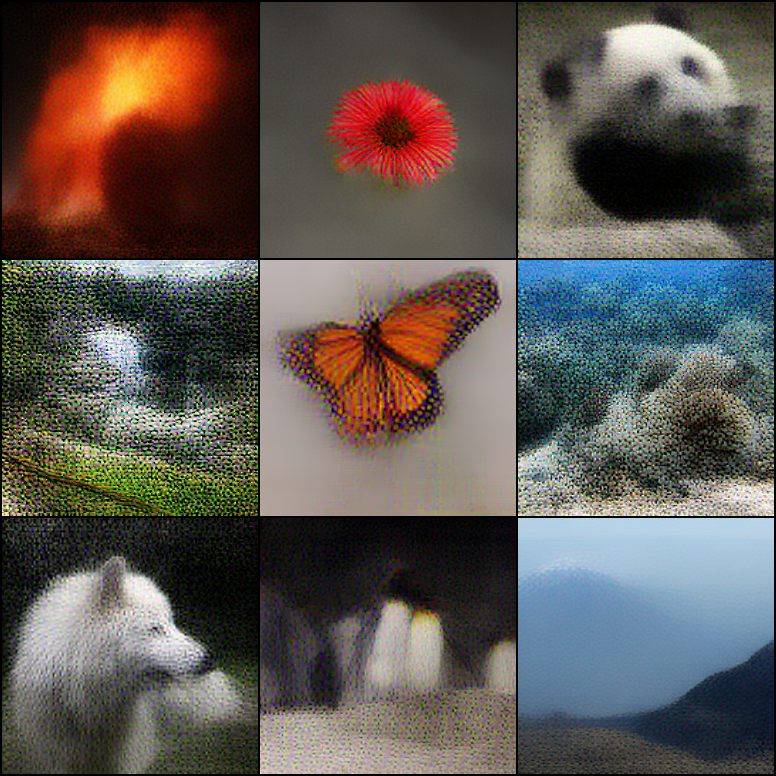}
    \caption*{MeanFlow}
    \label{fig:re-fig1}
  \end{subfigure}
  \hfill
    \begin{subfigure}{0.32\linewidth }
    \includegraphics[width=\linewidth]{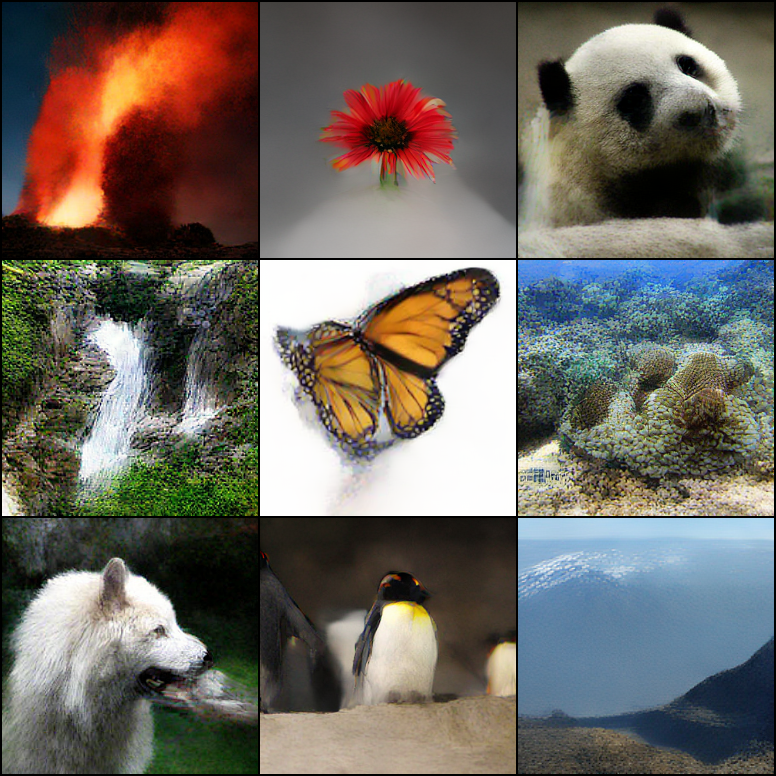}
    \caption*{sCMs}
    \label{fig:re-fig2}
  \end{subfigure}
  \hfill
    \begin{subfigure}{0.32\linewidth }
    \includegraphics[width=\linewidth]{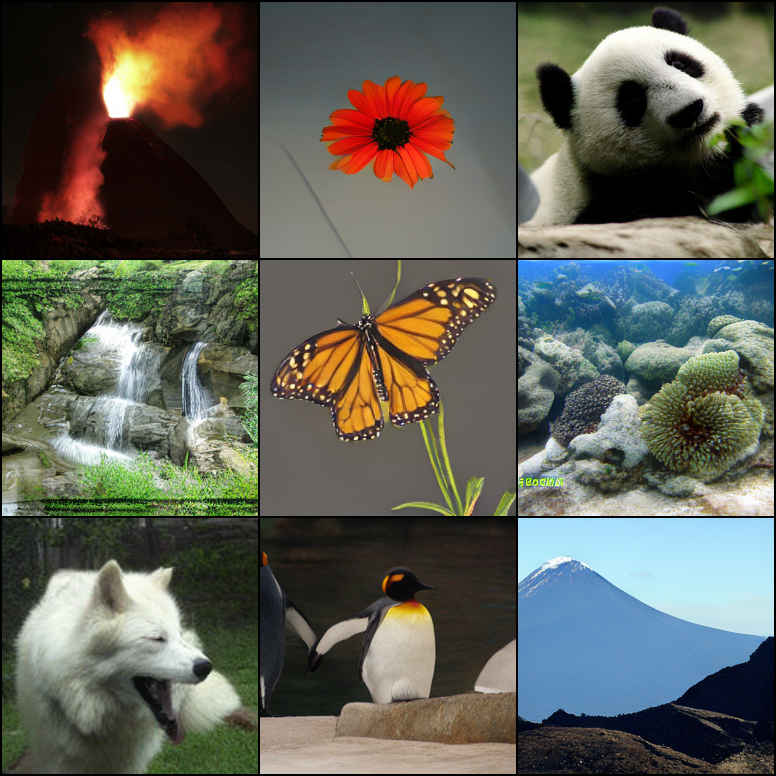}
    \caption*{Ours}
    \label{fig:re-fig3}
  \end{subfigure}
  \caption{Comparison of 1 NFE results under 8 GPUs resources. DE-CM demonstrated exceptional convergence efficiency at 16 GPU hours compared to sCMs and MeanFlow.}
  \label{fig:speed}
\end{figure}
\cref{fig:speed} presents 1 NFE visual comparisons under 16 GPU hours training.
Compared to MeanFlow, DE-CM yields clearer images, while compared to sCMs, it produces fewer artifacts.
This demonstrates the powerful training efficiency of DE-CM.
\cref{fig:grad_compare} (b) further illustrates the convergence performance of DE-CM under 8 GPUs resources, demonstrating strong efficiency and performance.

\subsection{Ablation Studies}
\cref{tab:ab-loss} presents ablation studies under different training objectives.
We can draw the following conclusions: 
1) The absence of CD loss prevents the achievement of few-step inference, as the distillation process is consistent-function-driven.
2) The absence of flow matching loss leads to suboptimal results due to unstable training. 
3) The absence of N2N loss affects multi-step sampling results due to the absence of initial noise perception.
We further investigate the impact of various hyperparameters and experimental configurations on the results. 
Covering all losses enables all steps to achieve significant gains.
Detailed implementation details are provided in Appendix \ref{sec:more_ab}.

\subsection{Discussion with Recent Efforts}

Our method is related to recent efforts on consistency-style training, but differs from them in both motivation and formulation.

\paragraph{\textbf{Discussion on stability.}}
Shortcut-style methods \cite{frans2024one} rely on flow matching (FM) to anchor the regime near $d=0$, making stability highly dependent on the model behavior as $d \to 0$. 
MeanFlow \cite{geng2025mean} implicitly includes FM through its flow-ratio formulation, and sLCT \cite{dao2025improved} also employs FM, though only in the low-noise regime. 
In contrast, DE-CM identifies the source of CMs instability by decomposing the continuous-time objective in \cref{eq:loss:two-loss}. 
From this perspective, FM acts as a left-boundary regularizer that offsets the instability introduced by the self-supervised term, rather than serving as the primary target of training. 
This explicit formulation also enables more flexible loss design, e.g., the additional $L_{\cos}$ term in \cref{eq:loss:fm}, and extends naturally to the full noise range.

\paragraph{\textbf{Further discussion on BOOT.}}
Our method is also related to BOOT \cite{gu2023boot}, since both optimize the segment from pure noise to $x_t$ along the PF-ODE trajectory. 
However, BOOT is proposed for {data-free distillation} through a bootstrapping mechanism based on a {discrete consistency property}, whereas our N2N trajectory (\cref{eq:loss:II loss}) is introduced to mitigate the error accumulation caused by long-jump CM sampling and is derived from the {continuous-time consistency model}. 
N2N therefore represents a theoretically grounded trajectory family within DE-CM. The full derivation is provided in \cref{ap:n2n}.

\begin{table}[!t]
\centering
\scriptsize
\caption{Comparison with recent methods. $\dagger$ indicates our repro.}
\vspace{-0.5em}
\begin{tabular}{lccccc}
\toprule
Method~~~~  & 1NFE-FID~~ & 2NFE-FID~~ & 1NFE-FID $\dagger$~~ & 2NFE-FID $\dagger$~~ \\
\midrule
AlphaFlow & 2.58 & 2.15 & 2.24 & 1.71 \\
CMT & 3.34 & - & 2.71 & 1.68 & \\
DE-CM & - & - & \textbf{1.70} & \textbf{1.33} \\
\bottomrule
\end{tabular}
\label{tab:releted}
\end{table}
\paragraph{\textbf{Comparison to recent efforts.}}
Recent methods \cite{geng2026improved, zhang2025alphaflow, hu2025cmt} improve training from different perspectives. 
iMF \cite{geng2026improved} reformulates MeanFlow as a $v$-prediction objective; 
Alphaflow \cite{zhang2025alphaflow} progressively adjusts $\alpha$ during training, transitioning from FM to shortcut-style training and eventually to MeanFlow; 
and CMT \cite{hu2025cmt} introduces an intermediate training stage to provide a more stable initialization. 
In contrast, DE-CM is derived from an objective-level diagnosis of CM instability and provides a corresponding correction mechanism. Quantitative comparisons are summarized in \cref{tab:releted}.

\begin{figure*}[!t]
  \centering
    \includegraphics[width=\linewidth]{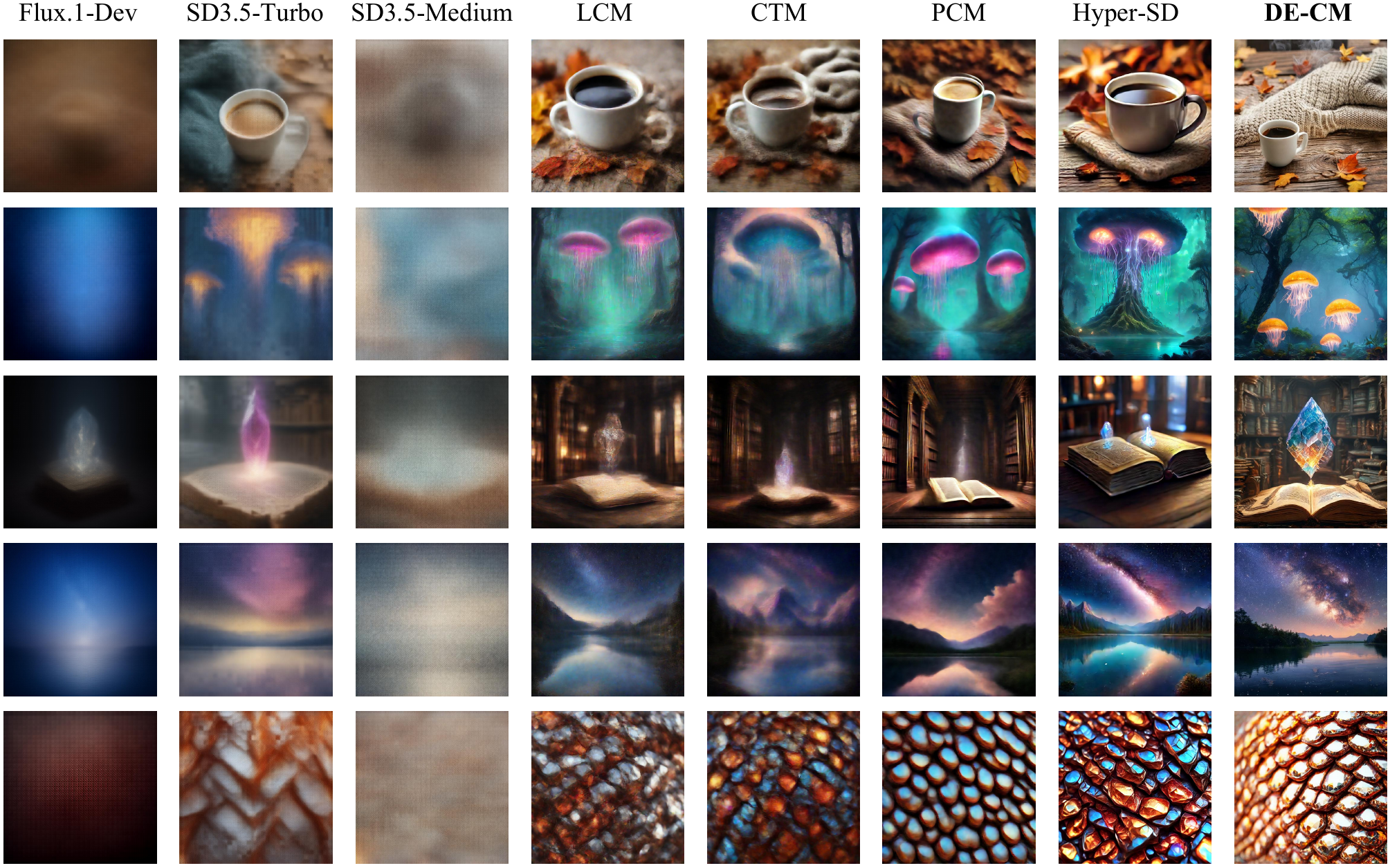}
   \caption{Visual quality comparison between DE-CM and existing methods under 1 NFE. DE-CM achieves superior naturalness in detail rendering, outperforming existing methods. The prompts used are provided in the Appendix \ref{ap:text_prompt}.
   }
   \label{fig:t2i-2}
\end{figure*}

\begin{table}[!t]
\centering
\caption{Ablation study on ImageNet-256 (FID $\downarrow$) analyzing the impact of training objective across different inference steps. We compared the model's performance under missing FM, CD, and N2N losses respectively.}
\label{tab:ab-loss}
\resizebox{\linewidth}{!}{
\begin{tabular}{lcccc}
\toprule
\textbf{Method} & ~~STEP=1 & ~~STEP=2 & ~~STEP=4 & ~~STEP=50 \\
\midrule
\multicolumn{4}{l}{\textbf{Training Objectives}} \\
Baseline Teacher~\cite{yao2025reconstruction}  & 309.51 & 230.79 & 111.59  & 2.77 \\
Baseline Teacher ($w = 1.75$) & 262.60 & 221.03 & 87.56 & 1.59 \\
\arrayrulecolor{black!40}\midrule
Flow Matching (FM) & 263.86 & 220.55 & 87.93 & 1.45 \\
CMs Distillation (CD) & 3.08 & 2.50 & 2.54  & 6.52 \\
FM + CD & 1.78 & 1.41 & 1.62 & 1.78 \\
FM + Noise-to-Noisy (N2N) Transmission  & 441.66 & 297.28 & 132.60 & 1.41 \\
FM + CD + N2N & \textbf{1.70} & \textbf{1.33} & \textbf{1.41} & \textbf{1.26} \\
\bottomrule
\end{tabular}
}
\end{table}

\section{Conclusion and Limitations}
In this work, we propose \textbf{Dual-End Consistency Model (DE-CM)}, a novel framework that effectively addresses the dual challenges of training instability and sampling inflexibility in CMs. 
DE-CM integrates continuous-time CMs objectives for efficient few-step distillation while employing flow matching as a boundary regularizer to ensure training stability. 
Furthermore, we introduce an innovative noise-to-noisy (N2N) mapping mechanism that enables flexible mapping from noise to any trajectory point, thus supporting to alleviate error accumulation. 
Extensive experiments demonstrate that our method sets a new state-of-the-art FID score of 1.70 for one-step generation on the ImageNet 256×256 benchmark with 250 epoch training.
Due to the incompatibility of the JVP operator with the FSDP framework and Flash Attention, the optimization process consumes a substantial amount of GPU memory, which limits the training of larger-scale models. 
Existing approaches try to addressed these problems, and it is hoped that more effective alternatives will emerge in the future.

\section*{Acknowledgments}
This work was supported by the National Natural Science Foundation of China under Grants No. 62293554 and U2336212; the ``Pioneer'' and ``Leading Goose'' R\&D Program of Zhejiang under Grants No. 2025C02022 and 2024C01161; the Zhejiang Provincial Natural Science Foundation of China under Grants No. LZ24F020002 and LD24F020007; the Ningbo Innovation ``Yongjiang 2035'' Key Research and Development Programme under Grant No. 2024Z292; and the Young Elite Scientists Sponsorship Program by CAST under Grant No. 2023QNRC001. The authors also gratefully acknowledge the support of the Zhejiang University Education Foundation Qizhen Scholar Foundation.

%
%
\bibliographystyle{splncs04}
\bibliography{main}

@String(ICML  = {Int. Conf. Mach. Learn.})

@String(ICLR  = {Int. Conf. Learn. Represent.})

@String(ICML  = {ICML})

@String(ICLR  = {ICLR})

@article{geng2025mean,
  title={Mean flows for one-step generative modeling},
  author={Geng, Zhengyang and Deng, Mingyang and Bai, Xingjian and Kolter, J Zico and He, Kaiming},
  journal={arXiv preprint arXiv:2505.13447},
  year={2025}
}

@article{peng2025flow,
  title={Flow-anchored consistency models},
  author={Peng, Yansong and Zhu, Kai and Liu, Yu and Wu, Pingyu and Li, Hebei and Sun, Xiaoyan and Wu, Feng},
  journal={arXiv preprint arXiv:2507.03738},
  year={2025}
}

@article{karras2022elucidating,
  title={Elucidating the design space of diffusion-based generative models},
  author={Karras, Tero and Aittala, Miika and Aila, Timo and Laine, Samuli},
  journal={Advances in neural information processing systems},
  volume={35},
  pages={26565--26577},
  year={2022}
}

@article{lu2024simplifying,
  title={Simplifying, stabilizing and scaling continuous-time consistency models},
  author={Lu, Cheng and Song, Yang},
  journal={arXiv preprint arXiv:2410.11081},
  year={2024}
}

@article{sabour2025align,
  title={Align Your Flow: Scaling Continuous-Time Flow Map Distillation},
  author={Sabour, Amirmojtaba and Fidler, Sanja and Kreis, Karsten},
  journal={arXiv preprint arXiv:2506.14603},
  year={2025}
}

@article{wu2025traflow,
  title={TraFlow: Trajectory Distillation on Pre-Trained Rectified Flow},
  author={Wu, Zhangkai and Fan, Xuhui and Wu, Hongyu and Cao, Longbing},
  journal={arXiv preprint arXiv:2502.16972},
  year={2025}
}

@article{zhou2025inductive,
  title={Inductive moment matching},
  author={Zhou, Linqi and Ermon, Stefano and Song, Jiaming},
  journal={arXiv preprint arXiv:2503.07565},
  year={2025}
}

@article{song2023improved,
  title={Improved techniques for training consistency models},
  author={Song, Yang and Dhariwal, Prafulla},
  journal={arXiv preprint arXiv:2310.14189},
  year={2023}
}

@article{geng2024consistency,
  title={Consistency models made easy},
  author={Geng, Zhengyang and Pokle, Ashwini and Luo, William and Lin, Justin and Kolter, J Zico},
  journal={arXiv preprint arXiv:2406.14548},
  year={2024}
}

@article{lipman2022flow,
  title={Flow matching for generative modeling},
  author={Lipman, Yaron and Chen, Ricky TQ and Ben-Hamu, Heli and Nickel, Maximilian and Le, Matt},
  journal={arXiv preprint arXiv:2210.02747},
  year={2022}
}

@article{liu2022flow,
  title={Flow straight and fast: Learning to generate and transfer data with rectified flow},
  author={Liu, Xingchao and Gong, Chengyue and Liu, Qiang},
  journal={arXiv preprint arXiv:2209.03003},
  year={2022}
}

@article{song2020denoising,
  title={Denoising diffusion implicit models},
  author={Song, Jiaming and Meng, Chenlin and Ermon, Stefano},
  journal={arXiv preprint arXiv:2010.02502},
  year={2020}
}

@article{lu2022dpm,
  title={Dpm-solver: A fast ode solver for diffusion probabilistic model sampling in around 10 steps},
  author={Lu, Cheng and Zhou, Yuhao and Bao, Fan and Chen, Jianfei and Li, Chongxuan and Zhu, Jun},
  journal={Advances in neural information processing systems},
  volume={35},
  pages={5775--5787},
  year={2022}
}

@article{lu2025dpm,
  title={Dpm-solver++: Fast solver for guided sampling of diffusion probabilistic models},
  author={Lu, Cheng and Zhou, Yuhao and Bao, Fan and Chen, Jianfei and Li, Chongxuan and Zhu, Jun},
  journal={Machine Intelligence Research},
  pages={1--22},
  year={2025},
  publisher={Springer}
}

@inproceedings{rombach2022high,
  title={High-resolution image synthesis with latent diffusion models},
  author={Rombach, Robin and Blattmann, Andreas and Lorenz, Dominik and Esser, Patrick and Ommer, Bj{\"o}rn},
  booktitle={Proceedings of the IEEE/CVF conference on computer vision and pattern recognition},
  pages={10684--10695},
  year={2022}
}

@article{ramesh2022hierarchical,
  title={Hierarchical text-conditional image generation with clip latents},
  author={Ramesh, Aditya and Dhariwal, Prafulla and Nichol, Alex and Chu, Casey and Chen, Mark},
  journal={arXiv preprint arXiv:2204.06125},
  volume={1},
  number={2},
  pages={3},
  year={2022}
}

@inproceedings{zhang2023adding,
  title={Adding conditional control to text-to-image diffusion models},
  author={Zhang, Lvmin and Rao, Anyi and Agrawala, Maneesh},
  booktitle={Proceedings of the IEEE/CVF international conference on computer vision},
  pages={3836--3847},
  year={2023}
}

@article{poole2022dreamfusion,
  title={Dreamfusion: Text-to-3d using 2d diffusion},
  author={Poole, Ben and Jain, Ajay and Barron, Jonathan T and Mildenhall, Ben},
  journal={arXiv preprint arXiv:2209.14988},
  year={2022}
}

@article{wang2023prolificdreamer,
  title={Prolificdreamer: High-fidelity and diverse text-to-3d generation with variational score distillation},
  author={Wang, Zhengyi and Lu, Cheng and Wang, Yikai and Bao, Fan and Li, Chongxuan and Su, Hang and Zhu, Jun},
  journal={Advances in neural information processing systems},
  volume={36},
  pages={8406--8441},
  year={2023}
}

@inproceedings{liu2023zero,
  title={Zero-1-to-3: Zero-shot one image to 3d object},
  author={Liu, Ruoshi and Wu, Rundi and Van Hoorick, Basile and Tokmakov, Pavel and Zakharov, Sergey and Vondrick, Carl},
  booktitle={Proceedings of the IEEE/CVF international conference on computer vision},
  pages={9298--9309},
  year={2023}
}

@article{liu2023audioldm,
  title={Audioldm: Text-to-audio generation with latent diffusion models},
  author={Liu, Haohe and Chen, Zehua and Yuan, Yi and Mei, Xinhao and Liu, Xubo and Mandic, Danilo and Wang, Wenwu and Plumbley, Mark D},
  journal={arXiv preprint arXiv:2301.12503},
  year={2023}
}

@inproceedings{evans2024fast,
  title={Fast timing-conditioned latent audio diffusion},
  author={Evans, Zach and Carr, CJ and Taylor, Josiah and Hawley, Scott H and Pons, Jordi},
  booktitle={Forty-first International Conference on Machine Learning},
  year={2024}
}

@article{blattmann2023stable,
  title={Stable video diffusion: Scaling latent video diffusion models to large datasets},
  author={Blattmann, Andreas and Dockhorn, Tim and Kulal, Sumith and Mendelevitch, Daniel and Kilian, Maciej and Lorenz, Dominik and Levi, Yam and English, Zion and Voleti, Vikram and Letts, Adam and others},
  journal={arXiv preprint arXiv:2311.15127},
  year={2023}
}

@article{wan2025,
      title={Wan: Open and Advanced Large-Scale Video Generative Models}, 
      author={Team Wan and Ang Wang and Baole Ai and Bin Wen and Chaojie Mao and Chen-Wei Xie and Di Chen and Feiwu Yu and Haiming Zhao and Jianxiao Yang and Jianyuan Zeng and Jiayu Wang and Jingfeng Zhang and Jingren Zhou and Jinkai Wang and Jixuan Chen and Kai Zhu and Kang Zhao and Keyu Yan and Lianghua Huang and Mengyang Feng and Ningyi Zhang and Pandeng Li and Pingyu Wu and Ruihang Chu and Ruili Feng and Shiwei Zhang and Siyang Sun and Tao Fang and Tianxing Wang and Tianyi Gui and Tingyu Weng and Tong Shen and Wei Lin and Wei Wang and Wei Wang and Wenmeng Zhou and Wente Wang and Wenting Shen and Wenyuan Yu and Xianzhong Shi and Xiaoming Huang and Xin Xu and Yan Kou and Yangyu Lv and Yifei Li and Yijing Liu and Yiming Wang and Yingya Zhang and Yitong Huang and Yong Li and You Wu and Yu Liu and Yulin Pan and Yun Zheng and Yuntao Hong and Yupeng Shi and Yutong Feng and Zeyinzi Jiang and Zhen Han and Zhi-Fan Wu and Ziyu Liu},
      journal = {arXiv preprint arXiv:2503.20314},
      year={2025}
}

@article{ho2020denoising,
  title={Denoising diffusion probabilistic models},
  author={Ho, Jonathan and Jain, Ajay and Abbeel, Pieter},
  journal={Advances in neural information processing systems},
  volume={33},
  pages={6840--6851},
  year={2020}
}

@article{song2019generative,
  title={Generative modeling by estimating gradients of the data distribution},
  author={Song, Yang and Ermon, Stefano},
  journal={Advances in neural information processing systems},
  volume={32},
  year={2019}
}

@article{song2020score,
  title={Score-based generative modeling through stochastic differential equations},
  author={Song, Yang and Sohl-Dickstein, Jascha and Kingma, Diederik P and Kumar, Abhishek and Ermon, Stefano and Poole, Ben},
  journal={arXiv preprint arXiv:2011.13456},
  year={2020}
}

@inproceedings{sohl2015deep,
  title={Deep unsupervised learning using nonequilibrium thermodynamics},
  author={Sohl-Dickstein, Jascha and Weiss, Eric and Maheswaranathan, Niru and Ganguli, Surya},
  booktitle={International conference on machine learning},
  pages={2256--2265},
  year={2015},
  organization={pmlr}
}

@inproceedings{esser2024scaling,
  title={Scaling rectified flow transformers for high-resolution image synthesis},
  author={Esser, Patrick and Kulal, Sumith and Blattmann, Andreas and Entezari, Rahim and M{\"u}ller, Jonas and Saini, Harry and Levi, Yam and Lorenz, Dominik and Sauer, Axel and Boesel, Frederic and others},
  booktitle={Forty-first international conference on machine learning},
  year={2024}
}

@article{wang2024phased,
  title={Phased consistency models},
  author={Wang, Fu-Yun and Huang, Zhaoyang and Bergman, Alexander and Shen, Dazhong and Gao, Peng and Lingelbach, Michael and Sun, Keqiang and Bian, Weikang and Song, Guanglu and Liu, Yu and others},
  journal={Advances in neural information processing systems},
  volume={37},
  pages={83951--84009},
  year={2024}
}

@inproceedings{yao2025reconstruction,
  title={Reconstruction vs. generation: Taming optimization dilemma in latent diffusion models},
  author={Yao, Jingfeng and Yang, Bin and Wang, Xinggang},
  booktitle={Proceedings of the Computer Vision and Pattern Recognition Conference},
  pages={15703--15712},
  year={2025}
}

@inproceedings{ma2024sit,
  title={Sit: Exploring flow and diffusion-based generative models with scalable interpolant transformers},
  author={Ma, Nanye and Goldstein, Mark and Albergo, Michael S and Boffi, Nicholas M and Vanden-Eijnden, Eric and Xie, Saining},
  booktitle={European Conference on Computer Vision},
  pages={23--40},
  year={2024},
  organization={Springer}
}

@article{tian2024visual,
  title={Visual autoregressive modeling: Scalable image generation via next-scale prediction},
  author={Tian, Keyu and Jiang, Yi and Yuan, Zehuan and Peng, Bingyue and Wang, Liwei},
  journal={Advances in neural information processing systems},
  volume={37},
  pages={84839--84865},
  year={2024}
}

@inproceedings{peebles2023scalable,
  title={Scalable diffusion models with transformers},
  author={Peebles, William and Xie, Saining},
  booktitle={Proceedings of the IEEE/CVF international conference on computer vision},
  pages={4195--4205},
  year={2023}
}

@article{kim2023consistency,
  title={Consistency trajectory models: Learning probability flow ode trajectory of diffusion},
  author={Kim, Dongjun and Lai, Chieh-Hsin and Liao, Wei-Hsiang and Murata, Naoki and Takida, Yuhta and Uesaka, Toshimitsu and He, Yutong and Mitsufuji, Yuki and Ermon, Stefano},
  journal={arXiv preprint arXiv:2310.02279},
  year={2023}
}

@inproceedings{sauer2024adversarial,
  title={Adversarial diffusion distillation},
  author={Sauer, Axel and Lorenz, Dominik and Blattmann, Andreas and Rombach, Robin},
  booktitle={European Conference on Computer Vision},
  pages={87--103},
  year={2024},
  organization={Springer}
}

@inproceedings{yin2024one,
  title={One-step diffusion with distribution matching distillation},
  author={Yin, Tianwei and Gharbi, Micha{\"e}l and Zhang, Richard and Shechtman, Eli and Durand, Fredo and Freeman, William T and Park, Taesung},
  booktitle={Proceedings of the IEEE/CVF conference on computer vision and pattern recognition},
  pages={6613--6623},
  year={2024}
}

@article{yin2024improved,
  title={Improved distribution matching distillation for fast image synthesis},
  author={Yin, Tianwei and Gharbi, Micha{\"e}l and Park, Taesung and Zhang, Richard and Shechtman, Eli and Durand, Fredo and Freeman, Bill},
  journal={Advances in neural information processing systems},
  volume={37},
  pages={47455--47487},
  year={2024}
}

@article{song2023consistency,
  title={Consistency models},
  author={Song, Yang and Dhariwal, Prafulla and Chen, Mark and Sutskever, Ilya},
  year={2023}
}

@article{guo2025splitmeanflow,
  title={Splitmeanflow: Interval splitting consistency in few-step generative modeling},
  author={Guo, Yi and Wang, Wei and Yuan, Zhihang and Cao, Rong and Chen, Kuan and Chen, Zhengyang and Huo, Yuanyuan and Zhang, Yang and Wang, Yuping and Liu, Shouda and others},
  journal={arXiv preprint arXiv:2507.16884},
  year={2025}
}

@article{boffi2024flow,
  title={Flow map matching},
  author={Boffi, Nicholas M and Albergo, Michael S and Vanden-Eijnden, Eric},
  journal={arXiv preprint arXiv:2406.07507},
  volume={2},
  number={3},
  pages={9},
  year={2024}
}

@inproceedings{deng2009imagenet,
  title={Imagenet: A large-scale hierarchical image database},
  author={Deng, Jia and Dong, Wei and Socher, Richard and Li, Li-Jia and Li, Kai and Fei-Fei, Li},
  booktitle={2009 IEEE conference on computer vision and pattern recognition},
  pages={248--255},
  year={2009},
  organization={Ieee}
}

@article{heusel2017gans,
  title={Gans trained by a two time-scale update rule converge to a local nash equilibrium},
  author={Heusel, Martin and Ramsauer, Hubert and Unterthiner, Thomas and Nessler, Bernhard and Hochreiter, Sepp},
  journal={Advances in neural information processing systems},
  volume={30},
  year={2017}
}

@article{loshchilov2017decoupled,
  title={Decoupled weight decay regularization},
  author={Loshchilov, Ilya and Hutter, Frank},
  journal={arXiv preprint arXiv:1711.05101},
  year={2017}
}

@article{hu2022lora,
  title={Lora: Low-rank adaptation of large language models.},
  author={Hu, Edward J and Shen, Yelong and Wallis, Phillip and Allen-Zhu, Zeyuan and Li, Yuanzhi and Wang, Shean and Wang, Lu and Chen, Weizhu and others},
  journal={ICLR},
  volume={1},
  number={2},
  pages={3},
  year={2022}
}

@article{ren2024hyper,
  title={Hyper-sd: Trajectory segmented consistency model for efficient image synthesis},
  author={Ren, Yuxi and Xia, Xin and Lu, Yanzuo and Zhang, Jiacheng and Wu, Jie and Xie, Pan and Wang, Xing and Xiao, Xuefeng},
  journal={Advances in Neural Information Processing Systems},
  volume={37},
  pages={117340--117362},
  year={2024}
}

@article{luo2023latent,
  title={Latent consistency models: Synthesizing high-resolution images with few-step inference},
  author={Luo, Simian and Tan, Yiqin and Huang, Longbo and Li, Jian and Zhao, Hang},
  journal={arXiv preprint arXiv:2310.04378},
  year={2023}
}

@misc{flux.1-dev2024,
  author       = {black-forest-labs},
  title        = {FLUX.1-dev},
  year         = {2024},
  howpublished = {\url{https://huggingface.co/black-forest-labs/FLUX.1-dev}},
}

@inproceedings{kang2023scaling,
  title={Scaling up gans for text-to-image synthesis},
  author={Kang, Minguk and Zhu, Jun-Yan and Zhang, Richard and Park, Jaesik and Shechtman, Eli and Paris, Sylvain and Park, Taesung},
  booktitle={Proceedings of the IEEE/CVF conference on computer vision and pattern recognition},
  pages={10124--10134},
  year={2023}
}

@inproceedings{sauer2022stylegan,
  title={Stylegan-xl: Scaling stylegan to large diverse datasets},
  author={Sauer, Axel and Schwarz, Katja and Geiger, Andreas},
  booktitle={ACM SIGGRAPH 2022 conference proceedings},
  pages={1--10},
  year={2022}
}

@inproceedings{esser2021taming,
  title={Taming transformers for high-resolution image synthesis},
  author={Esser, Patrick and Rombach, Robin and Ommer, Bjorn},
  booktitle={Proceedings of the IEEE/CVF conference on computer vision and pattern recognition},
  pages={12873--12883},
  year={2021}
}

@article{dhariwal2021diffusion,
  title={Diffusion models beat gans on image synthesis},
  author={Dhariwal, Prafulla and Nichol, Alexander},
  journal={Advances in neural information processing systems},
  volume={34},
  pages={8780--8794},
  year={2021}
}

@inproceedings{chang2022maskgit,
  title={Maskgit: Masked generative image transformer},
  author={Chang, Huiwen and Zhang, Han and Jiang, Lu and Liu, Ce and Freeman, William T},
  booktitle={Proceedings of the IEEE/CVF conference on computer vision and pattern recognition},
  pages={11315--11325},
  year={2022}
}

@article{frans2024one,
  title={One step diffusion via shortcut models},
  author={Frans, Kevin and Hafner, Danijar and Levine, Sergey and Abbeel, Pieter},
  journal={arXiv preprint arXiv:2410.12557},
  year={2024}
}

@misc{jackyhatetexttoimag2M,
  title={Text-to-image-2M},
  author={Jackyhate},
  year={2025},
  howpublished = {\url{https://huggingface.co/datasets/jackyhate/text-to-image-2M}},
  doi={10.57967/hf/3066}
}

@inproceedings{xu2023imagereward,
  title={ImageReward: learning and evaluating human preferences for text-to-image generation},
  author={Xu, Jiazheng and Liu, Xiao and Wu, Yuchen and Tong, Yuxuan and Li, Qinkai and Ding, Ming and Tang, Jie and Dong, Yuxiao},
  booktitle={Proceedings of the 37th International Conference on Neural Information Processing Systems},
  pages={15903--15935},
  year={2023}
}

@inproceedings{radford2021learning,
  title={Learning transferable visual models from natural language supervision},
  author={Radford, Alec and Kim, Jong Wook and Hallacy, Chris and Ramesh, Aditya and Goh, Gabriel and Agarwal, Sandhini and Sastry, Girish and Askell, Amanda and Mishkin, Pamela and Clark, Jack and others},
  booktitle={International conference on machine learning},
  pages={8748--8763},
  year={2021},
  organization={PmLR}
}

@inproceedings{li2022blip,
  title={Blip: Bootstrapping language-image pre-training for unified vision-language understanding and generation},
  author={Li, Junnan and Li, Dongxu and Xiong, Caiming and Hoi, Steven},
  booktitle={International conference on machine learning},
  pages={12888--12900},
  year={2022},
  organization={PMLR}
}

@article{dao2025improved,
  title={Improved training technique for latent consistency models},
  author={Dao, Quan and Doan, Khanh and Liu, Di and Le, Trung and Metaxas, Dimitris},
  journal={arXiv preprint arXiv:2502.01441},
  year={2025}
}

@inproceedings{gu2023boot,
  title={Boot: Data-free distillation of denoising diffusion models with bootstrapping},
  author={Gu, Jiatao and Zhai, Shuangfei and Zhang, Yizhe and Liu, Lingjie and Susskind, Joshua M},
  booktitle={ICML 2023 Workshop on Structured Probabilistic Inference $\{$$\backslash$\&$\}$ Generative Modeling},
  year={2023}
}

@inproceedings{geng2026improved,
  title={Improved mean flows: On the challenges of fastforward generative models},
  author={Geng, Zhengyang and Lu, Yiyang and Wu, Zongze and Shechtman, Eli and Kolter, J Zico and He, Kaiming},
  booktitle={Proceedings of the IEEE/CVF Conference on Computer Vision and Pattern Recognition},
  pages={30467--30476},
  year={2026}
}

@article{zhang2025alphaflow,
  title={Alphaflow: Understanding and improving meanflow models},
  author={Zhang, Huijie and Siarohin, Aliaksandr and Menapace, Willi and Vasilkovsky, Michael and Tulyakov, Sergey and Qu, Qing and Skorokhodov, Ivan},
  journal={arXiv preprint arXiv:2510.20771},
  year={2025}
}

@article{hu2025cmt,
  title={Cmt: Mid-training for efficient learning of consistency, mean flow, and flow map models},
  author={Hu, Zheyuan and Lai, Chieh-Hsin and Mitsufuji, Yuki and Ermon, Stefano},
  journal={arXiv preprint arXiv:2509.24526},
  year={2025}
}

\clearpage
\appendix
\counterwithin{figure}{section}
\counterwithin{table}{section}
\counterwithin{equation}{section}
\section*{Supplementary Material}

\section{Proofs and Derivations}
\label{sec:proof}
\subsection{Continuous-time CMs.}
Consistency model (CM) $f_{\theta}(x_t,t,s)$ is trained to map the noisy input $x_t$ directly to the corresponding clean
data $x_1$ in one step. 
The following discrete-time consistency loss function is defined as:
\begin{equation}
\begin{aligned}
& \mathbb{E}_{x_t,t,s|s=1} \left[ w(t, s) d (f_{\theta} (x_t, t, s) - f_{\theta^-} (x_{t^{\prime}}, t^{\prime}, s) )  \right],
\label{ap:eq:dis-t}
\end{aligned}
\end{equation}
where $\theta^-$ denotes $stopgrad(\theta)$, $t,s,t^\prime$ are timesteps, $t^\prime > t$ and $w(t,s)$ is the weighting
function designed for stable training. 
Let $t$ and $t^{\prime} = t + \epsilon $ denote two adjacent starting timesteps for a small $\epsilon > 0$, $x_{t^\prime}$ is obtained by $x_{t^\prime} = x_t + (t^\prime - t) v_\phi(x_t,t)$ which means using a 1-step Euler ODE Solver along the PF-ODE trajectory at the noisy point $x_t$ and $d(\cdot,\cdot)$ denotes $L_2$ norm.
Thus \cref{ap:eq:dis-t}'s gradient with respect to $\theta$ converges to:
\begin{equation}
\begin{aligned}
    \nabla_{\theta}\mathbb{E}_{x_{t},t,s|s=1} \left[ -w^\prime(t,s) f_{\theta}^{\top}(x_{t},t,s) \frac{\mathrm{d} f_{\theta^-}(x_{t},t,s)}{\mathrm{d}t} \right],
\label{ap:eq:con-t}
\end{aligned}
\end{equation}
when $\epsilon \rightarrow 0$. The derivation is as follows:
\begin{equation}
\begin{aligned}
& \nabla_\theta  \mathbb{E}_{{x}_t,t,s|s=1}\left[w(t,s)\|{f}_\theta({x}_t,t,s)-{f}_{\theta^-}({x}_{t^{\prime}},t^{\prime},s)\|_2^2\right] \\
& =   
\mathbb{E}_{{x}_t,t,s|s=1}\left[ 2 \cdot w(t,s)\nabla_\theta{f}^\top_\theta({x}_t,t,s) \right. 
\left. ({f}_{\theta^-}({x}_t,t,s)- 
{f}_{\theta^-}({x}_{t^{\prime}}, t^{\prime},s)\right)] \\
& = \mathbb{E}_{{x}_t,t,s|s=1}\bigg[ 2 \cdot w(t,s)\nabla_\theta{f}^\top_\theta({x}_t,t,s) \bigg({f}_{\theta^-}({x}_t,t,s) -  ({f}_{\theta^-}({x}_t,t,s) \\
& ~~~~~~~~~~~~~~~~~~~~~~~~~~~~+\frac{\partial{f}_{\theta^-}}{\partial{x}}({x}_{t^{\prime}}-{x}_t) + \frac{\partial{f}_{\theta^-}}{\partial t}(t^{\prime}-t)+O_{\epsilon^2}) \bigg) \bigg] \\
 & = \mathbb{E}_{{x}_t,t,s=1} \bigg[ 2 \cdot w(t,s)\nabla_\theta{f}^\top_\theta({x}_t,t,s) \cdot  \\ 
 & \left(-\left(\frac{\partial{f}_{\theta^-}}{\partial{x}}   (x_t + (t^\prime-t)\cdot{v}_\phi({x}_t,t)   
 - x_t)+\frac{\partial{f}_{\theta^-}}{\partial t}(t^\prime-t)\right)\right)\bigg]+O_{\epsilon} \\
 & =
 \mathbb{E}_{{x}_t,t,s=1}\left[-2 \cdot w(t,s)\nabla_\theta{f}^\top_\theta({x}_t,t,s)\left(\frac{\partial{f}_{\theta^-}}{\partial{x}} \cdot \right. \right.  \\
 &   ~~~~~~~~~~~~~~~~~~~~~~~~ \left. \left. (t^\prime-t)\cdot{v}_\phi({x}_t,t)+\frac{\partial{f}_{\theta^-}}{\partial t}(t^\prime-t)\right)\right]+O_{\epsilon} \\
 & =
 \mathbb{E}_{{x}_t,t,s=1}\bigg[-2 \cdot \epsilon  w(t,s) \cdot \nabla_\theta{f}^\top_\theta({x}_t,t,s)  \bigg(\frac{\partial{f}_{\theta-}}{\partial{x}}{v}_\phi({x}_t,t)+\frac{\partial{f}_{\theta-}}{\partial t}\bigg)\bigg]+O_{\epsilon} \\
 & \approx \nabla_\theta \mathbb{E}_{{x}_t,t,s|s=1}\left[-w^\prime(t,s){f}^\top_\theta({x}_t,t,s)\left(\frac{\mathrm{d}{f}_{\theta^-}({x}_t,t,s)}{\mathrm{d}t}\right)\right]
\end{aligned}
\label{ap:loss:der-cm}
\end{equation}
where $w^\prime (t,s) = 2w(t,s)\epsilon$. In \cref{ap:loss:der-cm}, only $\nabla_\theta{f}_\theta({x}_t,t,s)$ requires calculating its gradient with respect to $\theta$. Considering that $f_\theta({x}_t,t,s)$ is in the form of general flow maps, $f_\theta({x}_t,t,s) = x_t + (s-t)F_\theta(x_t,t,s)$ and $F_\theta$ is a neural network,
we can obtain
\begin{equation}
\begin{aligned}
& \mathbb{E}_{{x}_t,t,s|s=1}\left[-w^\prime(t,s)\nabla_\theta{f}^\top_\theta({x}_t,t,s)\left(\frac{\mathrm{d}{f}_{\theta^-}({x}_t,t,s)}{\mathrm{d}t}\right)\right] \\
= &
\mathbb{E}_{{x}_t,t,s|s=1}\bigg[-w^\prime(t,s)\nabla_\theta(x_t + (s-t){F}^\top_\theta({x}_t,t,s)) \cdot \\
&~~~~~~~~~~~~~~~~~~~~~~~~~~~~~~~~~~~  \bigg(\frac{\mathrm{d}(x_t + (s-t){F}_{\theta^-}({x}_t,t,s))}{\mathrm{d}t}\bigg)\bigg] \\
= & 
\mathbb{E}_{{x}_t,t,s|s=1}\bigg[-w^\prime(t,s) (s-t) \nabla_\theta{F}^\top_\theta({x}_t,t,s) \cdot \bigg(\frac{\mathrm{d}x_t}{\mathrm{d}t}  \\
& ~~~~~~~~~~~~~~~~~ + (s-t)\frac{\mathrm{d}{F}_{\theta^-}({x}_t,t,s)}{\mathrm{d}t} - {F}_{\theta^-}({x}_t,t,s)\bigg)\bigg] \\
= & 
\mathbb{E}_{{x}_t,t,s|s=1}\bigg[-w^\prime(t,s) (s-t) \nabla_\theta{F}^\top_\theta({x}_t,t,s) \cdot  \bigg( v_\phi (x_t,t) \\
& ~~~~~~~~~~~~~~~ + (s-t)(\frac{\partial {F}_{\theta^-}}{\partial x_t} \frac{\partial x_t}{\partial t} + \frac{\partial {F}_{\theta^-}}{\partial t} \frac{\partial t}{\partial t}  + \frac{\partial {F}_{\theta^-}}{\partial s} \frac{\partial s}{\partial t}) - {F}_{\theta^-}({x}_t,t,s) \bigg)\bigg] \\
= & 
\mathbb{E}_{{x}_t,t,s|s=1}\bigg[-w^\prime(t,s) (s-t) \nabla_\theta{F}^\top_\theta({x}_t,t,s) \cdot   \\
&  \bigg( \underbrace{v_\phi (x_t,t)+ (s-t)(\frac{\partial {F}_{\theta^-}}{\partial x_t}v_\phi + \frac{\partial {F}_{\theta^-}}{\partial t} - {F}_{\theta^-}({x}_t,t,s)}_{\text{gradient} ~~ g} \bigg)\bigg], \\
\end{aligned}
\label{ap:eq:loss:tail}
\end{equation}
\cref{ap:eq:loss:tail} demonstrates that the training objective is constructed from the reference velocity and the neural network's gradient at time t.
Considering $\nabla_\theta F_\theta \cdot y = \frac{1}{2} \nabla_\theta \| F_\theta - F_{\theta^-} + y \|^2_2$, we can formulate the training objectives as follows:
\begin{equation}
\begin{aligned}
 &   \mathcal{L}_\text{cm} = \mathbb{E}_{{x}_t,t,s|s=1}\bigg[ \frac{1}{2} \bigg\|{F}_\theta({x}_t,t,s) - {F}_{\theta^-}({x}_t,t,s) -  \\
& ~~~~~~~ w^\prime(t,s) \bigg( v_\phi (x_t,t) + (s-t)(\frac{\partial {F}_{\theta^-}}{\partial x_t}  v_\phi (x_t,t) + \frac{\partial {F}_{\theta^-}}{\partial t}  ) - {F}_{\theta^-}({x}_t,t,s)\bigg) \bigg \|^2_2 \bigg] \\
& =   
\mathbb{E}_{{x}_t,t,s|s=1}\bigg[ \frac{1}{2} \bigg\|{F}_\theta({x}_t,t,s) - (1 - w^\prime(t,s)){F}_{\theta^-}({x}_t,t,s)   \\
& - w^\prime(t,s) \bigg( v_\phi (x_t,t) + (s-t)(\frac{\partial {F}_{\theta^-}}{\partial x_t}  v_\phi (x_t,t) + \frac{\partial {F}_{\theta^-}}{\partial t}  \bigg)  \bigg\|^2_2 \bigg], \\
\end{aligned}
\label{ap:eq:loss:cm-final}
\end{equation}
here we incorporate the term $(s-t)$ into the weighting $w^\prime (t,s)$.  
Thus the final weighting denotes $w^\prime (t,s) = 2w(t,s) \cdot \epsilon \cdot (s-t)$
Generally, researchers set the self-designed weighting function $w(t, s)$ to $\frac{1}{\epsilon}$ to achieve faster convergence. 
However, this approach also leads to more unstable training when no additional constraints are applied.

\subsection{Noise-to-Noisy Mapping}
\label{ap:n2n}
Taking the timestep $t$ as the right endpoint, while the left endpoint $r$ approaches $0$. The objective of our discrete-time loss function is to minimize:
\begin{equation}
    \mathbb{E}_{x_r,r,t|r \rightarrow 0} \left[ w(r,t) d( f_\theta(x_r, r, t), \mathcal{S_\psi}(f_{\theta^-}(x_r, r, t^\prime))  )  \right],
    \label{ap:eq:dis-cm-right}
\end{equation}
where $\theta^-$ denotes $stopgrad(\theta)$, $r,t,t^\prime$ are timesteps, $t > t^\prime$ here and $w(r,t)$ is the custom weighting function. $\mathcal{S_\psi}$ denotes the Euler ODE Solver that predict $x_t$ from $x_t^\prime$ using one step.
Similar to \cref{ap:eq:dis-t}, the objective function guides the suboptimal outputs $f_\theta(x_r, r, t)$ towards the superior solution $\mathcal{S_\psi}(f_{\theta^-}(x_r, r, t^\prime))$.
Similarly, let $t$ and $t^{\prime} = t - \epsilon $ denote two adjacent starting timesteps for a small $\epsilon > 0$ and $d(\cdot , \cdot)$ denotes $L_2$ norm.
We can derive that
\begin{equation}
    \begin{aligned}
 & \nabla_\theta\mathbb{E}_{{x}_r,r \rightarrow 0,t}\left[w(r,t)\|{f}_\theta({x}_r,r,t)-\mathcal{S_\psi}(f_{\theta^-}(x_r, r, t^\prime))\|_2^2\right] \\
 & = \mathbb{E}_{{x}_r,r \rightarrow 0,t}\left[2 \cdot w(r,t)\nabla_\theta{f}^\top_\theta({x}_r,r,t)\left({f}_\theta({x}_r,r,t)
- \mathcal{S_\psi}(f_{\theta^-}(x_r, r, t^\prime)) \right)\right] \\
 & =  
 \mathbb{E}_{{x}_r,r \rightarrow 0,t}\left[2 \cdot w(r,t)\nabla_\theta{f}^\top_\theta({x}_r,r,t)\left({f}_\theta({x}_r,r,t)- \right. \right. \\
 & ~~~~~~~~~ \left. \left. \left({f}_{\theta^-}({x}_r,r,t^{\prime})+(t-t^{\prime})\cdot{v}_\phi({f}_{\theta^-}({x}_r,r,t^{\prime}),t^{\prime}))\right)\right]\right. \\
 & =
\mathbb{E}_{{x}_r,r \rightarrow 0,t}
\bigg[2 \cdot w(r,t)\nabla_\theta{f}^\top_\theta({x}_r,r,t)
\bigg({f}_\theta({x}_r,r,t) - \bigg({f}_{\theta^-}({x}_r,r,t)- \\
& ~~~~~~~~~~~~~~~~~~~
\frac{\partial{f}_{\theta^-}({x}_r,r,t)}{\partial t}(t-t^{\prime})+O_{\epsilon^2}  +(t-t^{\prime})\cdot({v}_\phi({f}_{\theta^-}({x}_r,r,t^\prime),t^\prime) \bigg) \bigg)\bigg] \\
 &=
\mathbb{E}_{{x}_r,r \rightarrow 0,t}
\bigg[ 2 \cdot w(r,t)\nabla_\theta{f}^\top_\theta({x}_r,r,t)\bigg(\frac{\partial{f}_{\theta^-}({x}_r,r,t)}{\partial t} \cdot \\
& ~~~~~~~~~~~~~~~~~~~~~~~ (t-t^{\prime})-(t-t^{\prime}){v}_\phi({f}_{\theta^-}({x}_r,r,t^{\prime}),t^{\prime})\bigg)\bigg]+O_{\epsilon} \\
 &=
\mathbb{E}_{{x}_r,r \rightarrow 0,t}
\bigg[ 2 \epsilon w(r,t)
\nabla_\theta{f}^\top_\theta({x}_r,r,t)\bigg(\frac{\mathrm{d}{f}_{\theta^-}({x}_r,r,t)}{\mathrm{d} t}  -{v}_\phi({f}_{\theta^-}({x}_r,r,t^{\prime}),t^{\prime})\bigg)\bigg]+O_{\epsilon} \\
 & \approx
\nabla_\theta \mathbb{E}_{{x}_r,r \rightarrow 0,t}
\bigg[  w^\prime(r,t)
{f}^\top_\theta({x}_r,r,t)\bigg(\frac{\mathrm{d}{f}_{\theta^-}({x}_r,r,t)}{\mathrm{d} t}  -{v}_\phi({f}_{\theta^-}({x}_r,r,t),t)\bigg)\bigg],
\end{aligned}
\label{ap:eq:loss:n2n}
\end{equation}
where $w^\prime (r,t) = 2w(r,t)\epsilon$ and $v_\phi$ denotes the teacher model.
Considering endpoint $r$, we can obtain its objective following \cref{ap:loss:der-cm}:
\begin{equation}
\begin{aligned}
    \nabla_{\theta}\mathbb{E}_{x_{r},r,t|r \rightarrow 0} \left[ -w^\prime(r,t) f_{\theta}^{\top}(x_{r},r,t) \frac{\mathrm{d} f_{\theta^-}(x_{r},r,t)}{\mathrm{d}r} \right],
\label{ap:eq:con-r}
\end{aligned}
\end{equation}
\cref{ap:eq:loss:n2n} and \cref{ap:eq:con-r} share a similar form and we can unify them using coefficients $\lambda$ and $\gamma$.
the training objectives can be unified as
\begin{equation}
\begin{aligned}
    & \nabla_{\theta}\mathbb{E}_{x_{r},r,t|r \rightarrow 0} 
    \left[ w^\prime(r,t) f_{\theta}^{\top}(x_{r},r,t) 
    (\lambda \frac{\mathrm{d} f_{\theta^-}(x_{r},r,t)}{\mathrm{d}t} - \right. \\
    & \left.
    \gamma \frac{\mathrm{d}f_{\theta^-}(x_r,r,t)}{\mathrm{d}r} - \lambda \cdot v_\phi(f_{\theta^-}(x_r,r,t),t)) 
    \right].
\end{aligned}
\label{ap:eq:unify}
\end{equation}
Considering that $f_\theta({x}_r,r,t) = x_r + (t-r)F_\theta(x_r,r,t)$ and $F_\theta$ is a neural network, we can derive that
\begin{equation}
\begin{aligned}
& \mathcal{L}_\text{n2n}  = 
    \nabla_{\theta}\mathbb{E}_{x_{r},r,t|r \rightarrow 0} 
    \bigg[ w^\prime(r,t) (x_r + (t-r)F_{\theta}^{\top}(x_{r},r,t) ) \bigg(
     \\
    & 
    \lambda \frac{\mathrm{d} (x_r + (t-r) F_{\theta^-}(x_{r},r,t))}{\mathrm{d}t} -  \gamma \frac{\mathrm{d}(x_r + (t-r) F_{\theta^-}(x_r,r,t)}{\mathrm{d}r}  \\
    & ~~~~~~~~~~~~~~~~~~~~~~~~~~~~~~~~~~~~~~~~~~~~~~~~~~~~ - \lambda \cdot v_\phi(f_{\theta^-}(x_r,r,t),t)\bigg) 
    \bigg] \\
    & = 
    \nabla_{\theta}\mathbb{E}_{x_{r},r,t|r \rightarrow 0} 
    \bigg[ w^\prime(r,t) (t-r)F_{\theta}^{\top}(x_{r},r,t) \bigg( \lambda (t-r) \cdot
     \\
    & ~~~~~~~~~~
    \frac{\mathrm{d}  F_{\theta^-}(x_{r},r,t)}{\mathrm{d} t} - \gamma v_\phi(x_r,r) - \gamma (t-r)  \frac{\mathrm{d}F_{\theta^-}(x_r,r,t)}{\mathrm{d}r}  \\
    & ~~~~~~~~
    - \lambda \cdot v_\phi(f_{\theta^-}(x_r,r,t),t) + (\lambda + \gamma) F_{\theta^-}(x_{r},r,t) \bigg) 
    \bigg] \\
    & = 
    \nabla_{\theta}\mathbb{E}_{x_{r},r,t|r \rightarrow 0} 
    \bigg[ w^\prime(r,t) (t-r)F_{\theta}^{\top}(x_{r},r,t) \bigg(  (\lambda + \gamma) \cdot
     \\
    & ~~~~~~~~~~
     F_{\theta^-}(x_{r},r,t) - \gamma v_\phi(x_r,r) - \lambda v_\phi(f_{\theta^-}(x_r,r,t),t) \\
    & ~~~~~~~~~ - (t - r)(\gamma \frac{\partial F_{\theta^-}}{\partial x_r}  \frac{\partial x_r}{\partial r} + \gamma \frac{\partial F_{\theta^-}}{\partial r} \frac{\partial r}{\partial r}  - \lambda \frac{\partial F_{\theta^-}}{\partial t} \frac{\partial t}{\partial t}) \bigg) 
    \bigg] \\
    & = 
    \nabla_{\theta}\mathbb{E}_{x_{r},r,t|r \rightarrow 0} 
    \bigg[ w^\prime(r,t) (t-r)F_{\theta}^{\top}(x_{r},r,t) \bigg(  (\lambda + \gamma) \cdot
     \\
    & ~~~~
     F_{\theta^-}(x_{r},r,t) - \gamma v_\phi(x_r,r) - \lambda  v_\phi(f_{\theta^-}(x_r,r,t),t) \\
    & ~~~~ - (t - r)( \frac{\partial F_{\theta^-}}{\partial x_r}   ,  \frac{\partial F_{\theta^-}}{\partial r} ,  \frac{\partial F_{\theta^-}}{\partial t} ) \cdot (\gamma v_\phi (x_r,r),  \gamma, -\lambda )\bigg) 
    \bigg] \\
    & = 
    \nabla_{\theta}\mathbb{E}_{x_{r},r,t|r \rightarrow 0} 
    \bigg[ w^\prime(r,t) (t-r)F_{\theta}^{\top}(x_{r},r,t) \bigg(  (\lambda + \gamma) \cdot
     \\
    & ~~~~
     F_{\theta^-}(x_{r},r,t) - \gamma v_\phi(x_r,r) - \lambda  v_\phi(f_{\theta^-}(x_r,r,t),t)  - (t - r) \dot F_{\theta^-} \bigg) 
    \bigg] ,
\end{aligned}
\label{ap:eq:unify-der}
\end{equation}
where $\dot F = ( \frac{\partial F_{\theta^-}}{\partial x_r}   ,  \frac{\partial F_{\theta^-}}{\partial r} ,  \frac{\partial F_{\theta^-}}{\partial t} ) \cdot (\gamma v_\phi (x_r,r),  \gamma, -\lambda ) $  is related to the current state of the neural network parameters $\theta$. 
The final loss can be expressed in a form similar to \cref{ap:eq:loss:cm-final}:
\begin{equation}
\begin{aligned}
 &   \mathcal{L}_\text{n2n} = \mathbb{E}_{{x}_r,r,t|r \rightarrow 0}\bigg[ \frac{1}{2} \bigg\|{F}_\theta({x}_r,r,t) - {F}_{\theta^-}({x}_r,r,t)   \\
& ~~~ + w^\prime(r,t)(t-r) \bigg(  (\lambda + \gamma) F_{\theta^-}(x_{r},r,t) - (t - r) \dot F_{\theta^-} \\
& ~~~ - \gamma v_\phi(x_r,r) - \lambda  v_\phi(f_{\theta^-}(x_r,r,t),t)  \bigg)  \bigg \|^2_2 \bigg]. 
\end{aligned}
\label{ap:eq:loss:n2n-final}
\end{equation}

\subsection{Rethinking the Objectives of CMs}
Starting from \cref{ap:eq:con-t}, we derive the training objective loss anew:
\begin{equation}
\begin{aligned}
& \mathbb{E}_{{x}_t,t,s|s=1}\left[-w^\prime(t,s)\nabla_\theta{f}^\top_\theta({x}_t,t,s)\left(\frac{\mathrm{d}{f}_{\theta^-}({x}_t,t,s)}{\mathrm{d}t}\right)\right] \\
= &
\mathbb{E}_{{x}_t,t,s|s=1}\bigg[-w^\prime(t,s)\nabla_\theta(x_t + (s-t){F}^\top _\theta({x}_t,t,s))\cdot \\
&~~~~~~~~~~~~~~~~~~~~~~~~~~~~~~~~~~~  \bigg(\frac{\mathrm{d}(x_t + (s-t){F}_{\theta^-}({x}_t,t,s))}{\mathrm{d}t}\bigg)\bigg] \\
= & 
\mathbb{E}_{{x}_t,t,s|s=1}\bigg[-w^\prime(t,s) (s-t) \nabla_\theta{F}^\top_\theta({x}_t,t,s) \cdot \bigg(\frac{\mathrm{d}x_t}{\mathrm{d}t}  \\
& ~~~~~~~~~~~~~~~~~ + (s-t)\frac{\mathrm{d}{F}_{\theta^-}({x}_t,t,s)}{\mathrm{d}t} - {F}_{\theta^-}({x}_t,t,s)\bigg)\bigg] \\
= & 
\mathbb{E}_{{x}_t,t,s|s=1}\bigg[ w^{\prime}(t,s) (s-t) \nabla_\theta{F}^\top_\theta({x}_t,t,s) \cdot \bigg( -\frac{\mathrm{d}x_t}{\mathrm{d}t} \\
&  ~~~~~~~~~~~~ + {F}_{\theta^-}({x}_t,t,s)\bigg)\bigg]   + \mathbb{E}_{{x}_t,t,s|s=1}\bigg[w^\prime(t,s) (s-t) \cdot \\
&  ~~~~~~~~~~~~ \nabla_\theta{F}^\top_\theta({x}_t,t,s) \cdot \bigg( -(s-t)\frac{\mathrm{d}{F}_{\theta^-}({x}_t,t,s)}{\mathrm{d}t} \bigg)\bigg] \\
= & 
\mathbb{E}_{{x}_t,t,s|s=1}\bigg[ \frac{1}{2}w^{\prime}(t,s) (s-t)  \bigg \| {F}_{\theta}({x}_t,t,s) - v_\phi (x_t,t) \bigg \|_2^2 \bigg]  \\
& ~~~~~~~~~~~~ + \mathbb{E}_{{x}_t,t,s|s=1}\bigg[\frac{1}{2}w^\prime(t,s) (s-t) \bigg \|{F}_{\theta}({x}_t,t,s) \\
& ~~~~~~~~~~~~ - {F}_{\theta^-}({x}_t,t,s) -(s-t)\frac{\mathrm{d}{F}_{\theta^-}({x}_t,t,s)}{\mathrm{d}t} \bigg\|_2^2 \bigg] \\
= & 
\mathbb{E}_{{x}_t,t,s|s=1}\bigg[ \frac{1}{2}w^{\prime}(t,s) (s-t)  \bigg \| {F}_{\theta}({x}_t,t,s) - v_\phi (x_t,t) \bigg \|_2^2 \bigg]  \\
& + \mathbb{E}_{{x}_t,t,s|s=1}\bigg[\frac{1}{2}w^\prime(t,s) (s-t) \bigg \|{F}_{\theta}({x}_t,t,s) - \tilde v   \bigg\|_2^2 \bigg],
\end{aligned}
\label{ap:eq:loss:rethink}
\end{equation}
where $\tilde v = {F}_{\theta^-}({x}_t,t,s)  + (s-t)\frac{\mathrm{d}{F}_{\theta^-}({x}_t,t,s)}{\mathrm{d}t} $.
The original training objective can be decomposed into two losses, where the first loss can be regarded as alignment with the reference velocity, while the second term can be viewed as a self-supervised loss.
However, the self-supervised loss critically depends on the network's gradient at timestep $t$. 
This reliance introduces instability, as the gradient can become volatile under non-convergent parameters, resulting in erratic parameter updates.
A simple improvement to the unstable approach is to increase the weight of the first term, the stability loss.
However, due to the constraint of $s = 1$, the model tends to neglect self-supervised learning when learning instantaneous velocity.
This necessitates introducing an additional temporal condition variable $s$ to prevent divergence between the two losses when relying solely on the single temporal condition $t$ during training.
Considering $s \rightarrow t$, the second self-supervised loss term tends toward zero.
The remaining loss term is the standard flow matching loss $\mathcal{L}_\text{fm}$:
\begin{equation}
    \begin{aligned}
        \mathbb{E}_{{x}_t,t,s|s \rightarrow t} [ w(t) \| {F}_{\theta}({x}_t,t,s) - v_\phi (x_t,t)  \|_2^2 ] .
    \end{aligned}
    \label{ap:eq:loss:fm}
\end{equation}
\cref{ap:eq:loss:fm} decouples the stable optimization of the instantaneous trajectories at $s=t$, thereby avoiding loss degradation and training instability on the CMs trajectories.

\section{Connections to Existing Methods}
\subsection{MeanFlow}
MeanFlow defines the average velocity as the displacement between two time steps $t$ and $s$ divided by the time interval. 
Formally, the average velocity $F_\theta(x_t,t,s)$ is:
\begin{equation}
    F_\theta(x_t,t,s) = \frac{1}{s-t} \int_t^s v(x_\tau, \tau) \mathrm{d} \tau.
\end{equation}
And then calculates the total derivative of $t$ with respect to both sides, yielding MeanFlow Identity:
\begin{equation}
    F_\theta(x_t,t,s) = v_\phi (x_t,t) + (s-t) \frac{\mathrm{d} F_\theta(x_t,t,s)}{\mathrm{d} t}.
\end{equation}
Thus the training objectives can be derived:
\begin{equation}
\mathbb{E}_{{x}_t,t,s} \bigg[ \bigg\| {F}_{\theta}({x}_t,t,s) - sg\bigg(v_\phi (x_t,t) + (s-t) \frac{\mathrm{d} F_\theta(x_t,t,s)}{\mathrm{d} t}\bigg) \bigg\|_2^2 \bigg] 
\label{ap:loss:meanflow}
\end{equation}
\cref{ap:loss:meanflow}
It can be seen that \cref{ap:loss:meanflow} and \cref{ap:eq:loss:cm-final} are formally equivalent when $ w^\prime (t,s)=1$.
Furthermore, MeanFlow covers all enumeration, requiring that sub-trajectories constructed for any arbitrary $t$ and $s$ in the search space $\{(t,s)| t < s\}$ must be optimized.
However, DE-CM imposes constraints on $t$ and $s$, focusing more on the critical trajectory paths.
\subsection{sCoT}
sCoT regularizes the gradient of function $f_\theta (x_t,t,s)$.
In detail, the velocity loss function can be defined as
\begin{equation}
 \mathbb{E}_{{x}_t,t,s} \bigg[ \bigg\|  \frac{\mathrm{d} f_\theta(x_t,t,s)}{\mathrm{d} s}  - (\hat x_1 - x_0) \bigg\|_2^2 \bigg] 
 \label{ap:eq:loss:scot}
\end{equation}
This objective can be derived from \cref{ap:eq:loss:n2n} with $w^\prime(t,s) = 1$ and $v_\phi$ replaced by the pre-trained pair data $(x_0, \hat x_1)$(drawn form reflow).
\cref{ap:eq:loss:scot} must be optimized in conjunction with the discrete-time CTM loss to achieve more stable and higher-quality generation.
In contrast, our method eliminates the need for auxiliary regularization losses in discrete-time.

\subsection{sCM}
sCM is an improvement over continuous-time CM designed to enhance model training stability.
Its training objective is similar to \cref{ap:eq:con-t}:
\begin{equation}
    \nabla_{\theta}\mathbb{E}_{x_{t},t} \left[ -w(t) \sigma_d sin(t) F_{\theta}^{\top}(\frac{x_{t}}{\sigma_d},t) \frac{\mathrm{d} f_{\theta^-}(x_{t},t)}{\mathrm{d}t} \right],
\end{equation}
The primary difference from \cref{ap:eq:con-t} lies in its consistency function adhering to the construction under TrigFlow:
\begin{equation}
    f_\theta(x_t,t) = cos(t)x_t - sin(t)\sigma_d F_\theta(\frac{x_t}{\sigma_d}, c_\text{noise}(t))
\end{equation}
When our model employs TrigFlow to construct the diffusion forward process, both exhibit identical forms on the CMs trajectory.
\section{More Training Details}
\subsection{R-Value Decay}
In DE-CM, the $r$ value is designed to approach zero.
However, we find that setting $r = 0$ directly during the early stages of training caused significant oscillations in the gradients.
Therefore, we design the model to gradually decay $r$ to zero during the early stages, rather than setting it to zero from the outset.
The decay of $r$ is designed to
\begin{equation}
    r = max(0, (1- \frac{i}{i_\text{max}}) \delta)
\end{equation}
where $\delta$ is the initial value of decay, $i$ represents the current global iteration count, and $i_\text{max}$ denotes the total number of decay iterations. 
In the experiment, we set $i_\text{max}$ to 20,000 and $delta$ to $0.1$.
\subsection{Velocity Normalization}
We use the velocity calculated from the CFG guidance as a reference in distillation:
\begin{equation}
     v_\phi^\text{cfg} = v_\phi^\text{cond} + (w_\text{cfg} - 1)(v_\phi^\text{cond} - v_\phi^\text{uncond}),
\end{equation}
where $w_\text{cfg}$ denotes CFG weights. 
We perform normalization on the term $(v_\phi^\text{cond} - v_\phi^\text{uncond})$  to ensure its distribution does not deviate from $v_\phi^\text{cond}$, following APG.
Given the scaling factor $\eta$, we can obtain the final velocity:
\begin{equation}
     v_\phi^\text{cfg} = v_\phi^\text{cond} + (w_\text{cfg} - 1) \cdot min(1, \frac{\eta}{\| \Delta v\|}) \cdot \Delta v,
\end{equation}
where $\Delta v = v_\phi^\text{cond} - v_\phi^\text{uncond} $.
This approach not only prevents over-saturation, but also indirectly stabilizes training, as the JVP vector is constrained by regularization.
To identify the optimal $\eta$, we evaluate the FID of the teacher model across different $\eta$ configurations. The $\eta$ yielding the best FID will be selected for distillation.

\subsection{Timestep scheduling.}
Prior works, such as DiT and SiT have shown that the distribution used to sample $t, s$ in training impacts the generation quality.
The sampling distribution should generally encompass a broad spectrum across the entire $[0,1]$ interval.
However, we find that along PF-ODE trajectories defined by the teacher model, the distillation process only needs to be carried out at specific timestep.
We simply divide the interval $[0,1]$ into $N$ segments and perform random sampling only at the endpoints of these segments.
Just using simple uniform distribution sampling can yield better results.
Here we have not altered the continuous-time nature of the model. We have merely adjusted the sampling method.


\subsection{Interval update policy.}
While the ideal approach would involve sequentially straightening the trajectories $[t,1]$ and then $[r,t]$ in an alternating optimization manner, this poses significant practical challenges. 
The computational overhead of such a strategy is substantial and would prohibitively increase training time.
Our simple approach is to introduce an update frequency $freq$, updating the noisy-to-noise loss  every $freq$ iterations.


\section{More Ablation Studies}
\label{sec:more_ab}
\begin{table}[t]
\centering
\caption{Ablation studies for hyperparameters $\lambda,\gamma,\delta$ and design choices.}
\label{tab:hyper_all}

\begin{minipage}{0.48\linewidth}
\centering
\caption*{(a) Ablation on $\lambda,\gamma,\delta$}
\resizebox{\linewidth}{!}{
\begin{tabular}{lcccc}
\toprule
\textbf{Hyperparameters} & NFE=1 & NFE=2 & NFE=4 & NFE=50 \\
\midrule
$\delta = 0$ & 1.74 & 1.37 & 1.46 & 1.33 \\
$\delta = 0.1$ (\textbf{Ours}) & \textbf{1.70} & \textbf{1.33} & \textbf{1.41} & \textbf{1.26} \\
$\delta = 0.2$ & 1.71 & 1.34 & 1.42 & 1.31 \\
\midrule
$\lambda = 0.5, \gamma = 1$ (\textbf{Ours}) & \textbf{1.70} & \textbf{1.33} & \textbf{1.41} & \textbf{1.26} \\
$\lambda = 0, \gamma = 1$ & 1.73 & 1.44 & 1.49 & 1.36 \\
$\lambda = 1, \gamma = 1$ & 1.79 & 1.39 & 1.51 & 1.41 \\
$\lambda = 0.5, \gamma = 0.5$ & 1.82 & 1.41 & 1.56 & 1.38 \\
\bottomrule
\end{tabular}
}
\end{minipage}
\hfill
\begin{minipage}{0.48\linewidth}
\centering
\caption*{(b) Ablation on design choices}
\resizebox{\linewidth}{!}{
\begin{tabular}{lcccc}
\toprule
\textbf{Design Choices} & NFE=1 & NFE=2 & NFE=4 & NFE=50 \\
\midrule
\multicolumn{4}{l}{\textbf{Velocity Normalization}} \\
W/ Velocity Normalization (\textbf{Ours})  & \textbf{1.70} & 1.33 & \textbf{1.41} & \textbf{1.26} \\
W/O Velocity Normalization & 1.75 & \textbf{1.30} & 1.43 & 1.35 \\
\midrule
\multicolumn{4}{l}{\textbf{Timestep Scheduling}} \\
uniform-with-N-seg (\textbf{Ours}) & \textbf{1.70} & \textbf{1.33} & \textbf{1.41} & \textbf{1.26} \\
arctan-norm & 1.74 & 1.36 & 1.62 & 1.45 \\
log-norm & 1.83 & 1.45 & 1.67 & 1.41 \\
\midrule
\multicolumn{4}{l}{\textbf{Frequency Hyperparameter}} \\
$freq=1$ & 1.72 & 1.48 & 1.77 & 1.45 \\
\textbf{$freq=3$} (\textbf{Ours}) & \textbf{1.70} & \textbf{1.33} & \textbf{1.41} & \textbf{1.26} \\
$freq=5$ & 1.74 & 1.34 & 1.52 & 1.37 \\
\bottomrule
\end{tabular}
}
\end{minipage}
\end{table}
We conduct ablation studies on the hyperparameters $\delta,\lambda, \gamma$, as shown in \cref{tab:hyper_all} (a).
When setting $\delta=0$, the r-value decay strategy is not employed, and we observe significant gradient fluctuations during the early training phase. 
Consequently, the final FID score is also affected.
Setting a larger $\delta$ has a relatively minor impact on the experimental results.
In the ablation experiments on hyperparameters $\lambda$ and $\gamma$, we find that a small $\gamma$ leads to poor model optimization, while setting $\lambda$ to zero or increasing it adversely affects model performance.
The results suggest that the $\gamma$ term governs a gradient component that is crucial for model optimization and the $\lambda$ term serves as an auxiliary function that imposes regular constraints on training.

Table \ref{tab:hyper_all} (b) demonstrates that instantaneous velocity normalization effectively prevents oversaturation in the distillation results and improves FID performance at both 1 NFE and 50 NFE.
For 2 and 4 NFEs, the results are comparable, likely due to the noise injection process counteracting oversaturation.
The table also compares different time schedule designs, revealing that restricting the sampling process to specific time intervals benefits the rapid distillation of high-quality models.
Finally, it illustrates the influence of frequency selection: excessively low frequencies lead to training instability in the early stages, while overly high frequencies result in poor few-step performance.

\begin{table}[!t]
\centering
\scriptsize
\caption{Comparison with GAN integration.}
\begin{tabular}{lccccc}
\toprule
Epoch~~~~ & Method~~~~ & 1-NFE$\downarrow$~~~~ & 2-NFE$\downarrow$~~~~ & 4-NFE$\downarrow$~~~~ & 50-NFE$\downarrow$~~~~ \\
\midrule
\multirow{2}{*}{100} & DE-CM & 1.84 & 1.38 & 1.47 & \textbf{1.31}  \\
                     & DE-CM + GAN & \textbf{1.80} & \textbf{1.36} & \textbf{1.46} & \textbf{1.31} \\
\multirow{2}{*}{200} & DE-CM & 1.74 & \textbf{1.34} & \textbf{1.43} &  \textbf{1.28}\\
                     & DE-CM + GAN & \textbf{1.72} & 1.44 & 1.54 & 1.46 \\
\bottomrule
\end{tabular}
\label{tab:gan}
\end{table}

\Cref{tab:gan} presents the comparison with GAN integration. 
We observe that incorporating GAN improves convergence in the early stage of training. 
As training proceeds, the 1-NFE FID continues to improve, whereas the performance at higher NFEs shows an inflection point around 160 epochs. We attribute this behavior to the fact that GAN tends to optimize toward a single mapping, which may constrain the DE-CM objective.

\clearpage
\section{More Qualitative Results}
\begin{figure*}[!t]
  \centering
    \includegraphics[width=\linewidth]{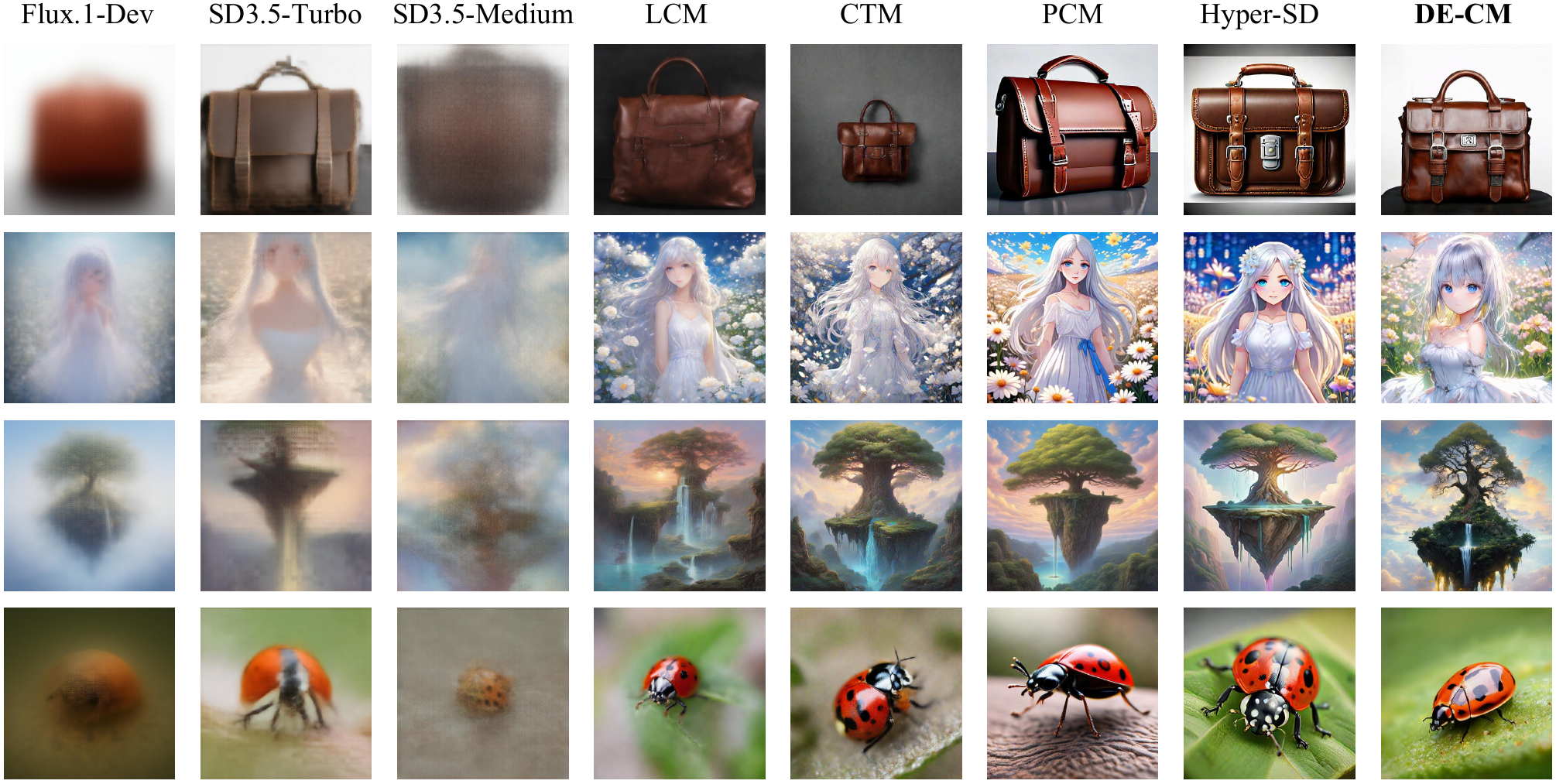}
   \caption{Visual quality comparison between DE-CM and existing methods under 2 NFE.
   }
   \label{fig:t2i-3}
\end{figure*}

\begin{figure*}[!t]
  \centering
    \includegraphics[width=\linewidth]{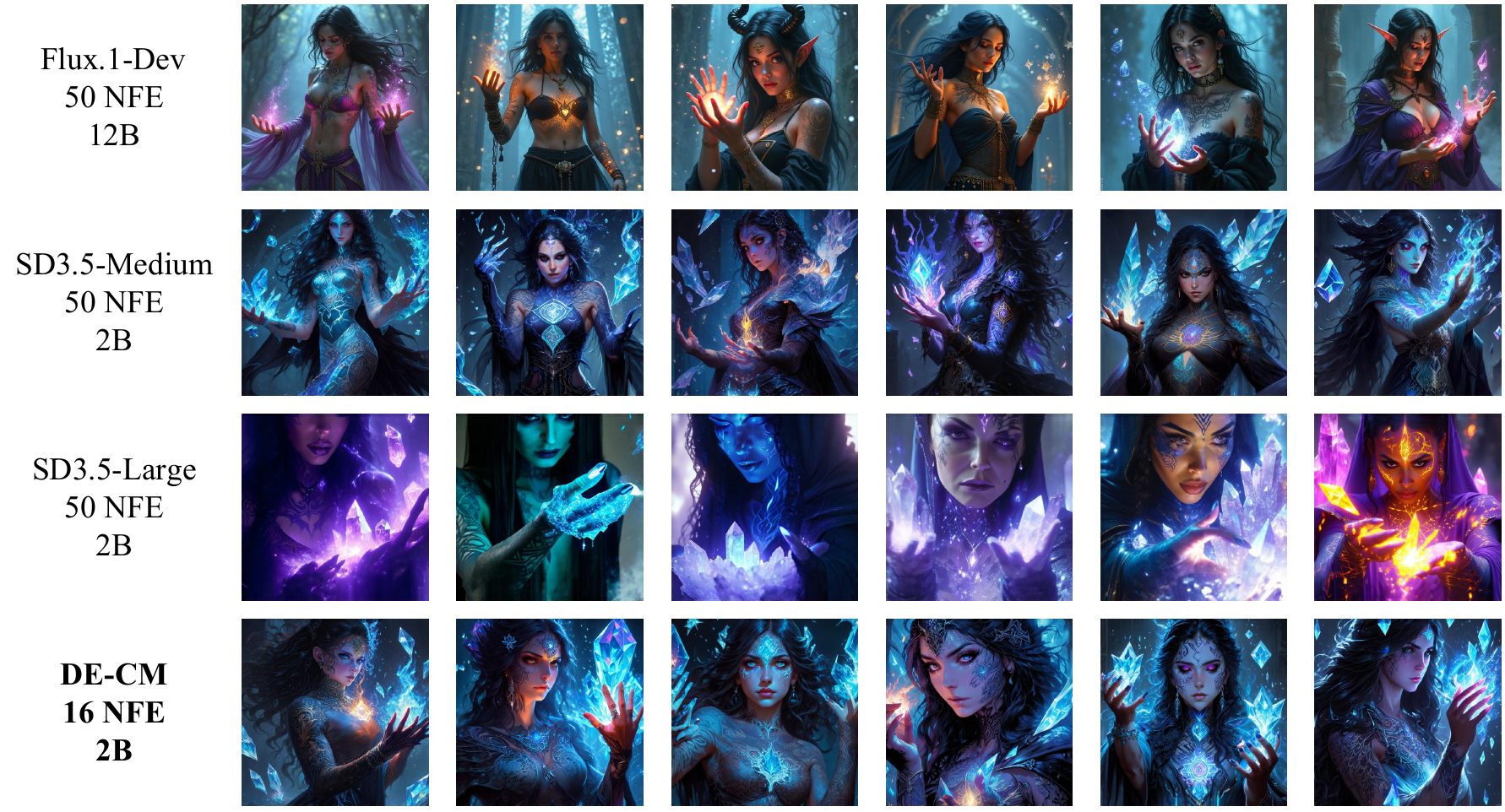}
   \caption{Visual comparison between DE-CM and large pre-trained models under high NFE.
   }
   \label{fig:t2i-4}
\end{figure*}

\begin{figure*}[!t]
  \centering
    \includegraphics[width=\linewidth]{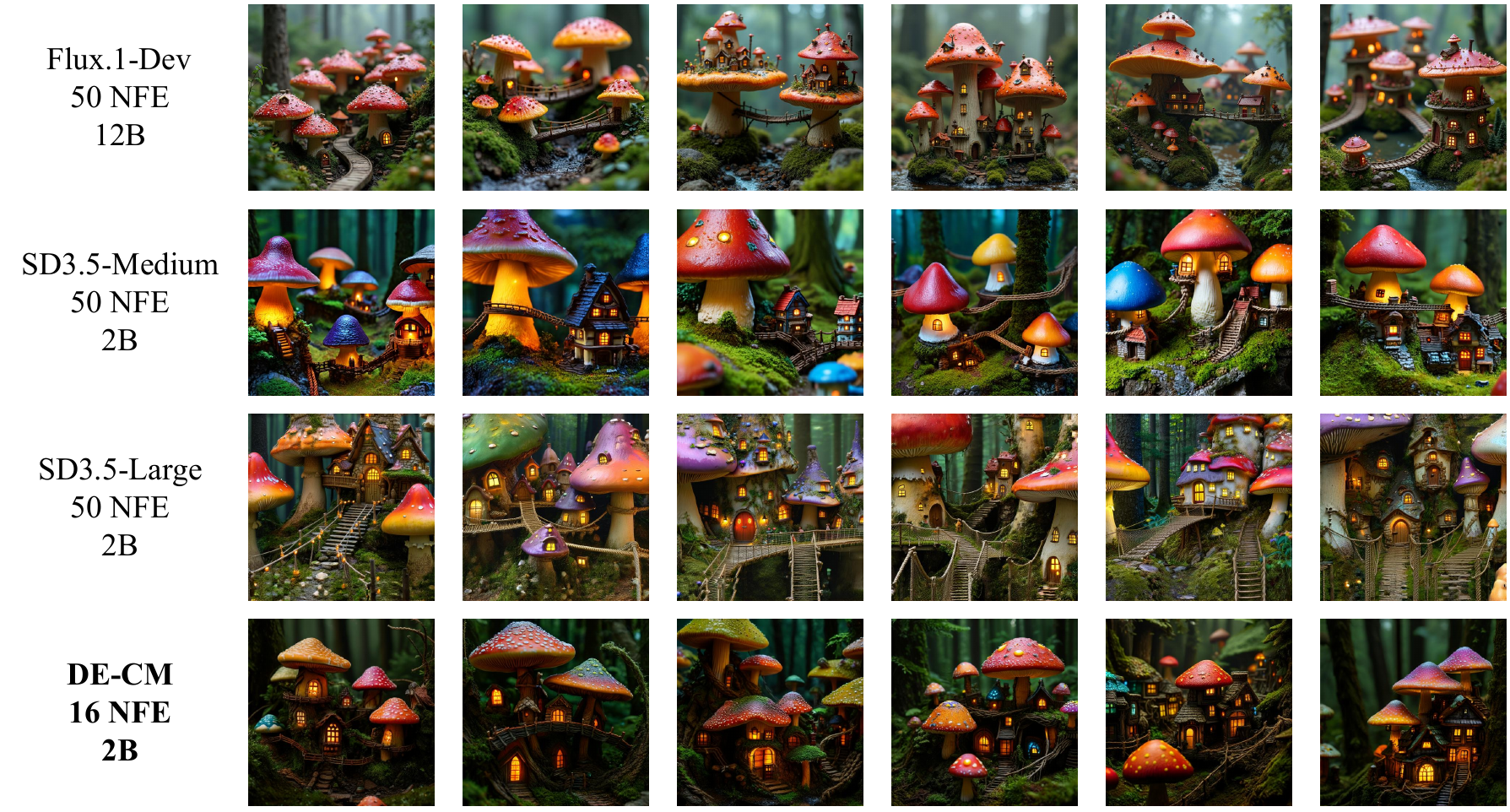}
   \caption{Visual comparison between DE-CM and large pre-trained models under high NFE.
   }
   \label{fig:t2i-5}
\end{figure*}

\begin{figure*}[!t]
  \centering
    \includegraphics[width=\linewidth]{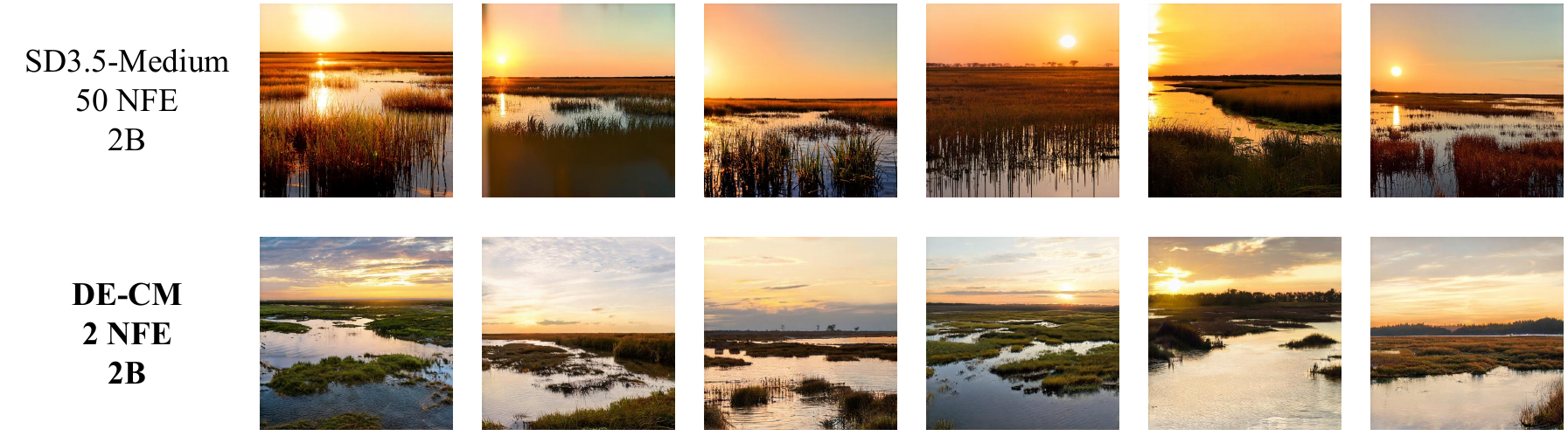}
   \caption{Visual comparison between DE-CM (2 NFE) and its distillation teacher SD3.5-Medium (50 NFE).
   }
   \label{fig:t2i-7}
\end{figure*}

\begin{figure*}[!t]
  \centering
    \includegraphics[width=\linewidth]{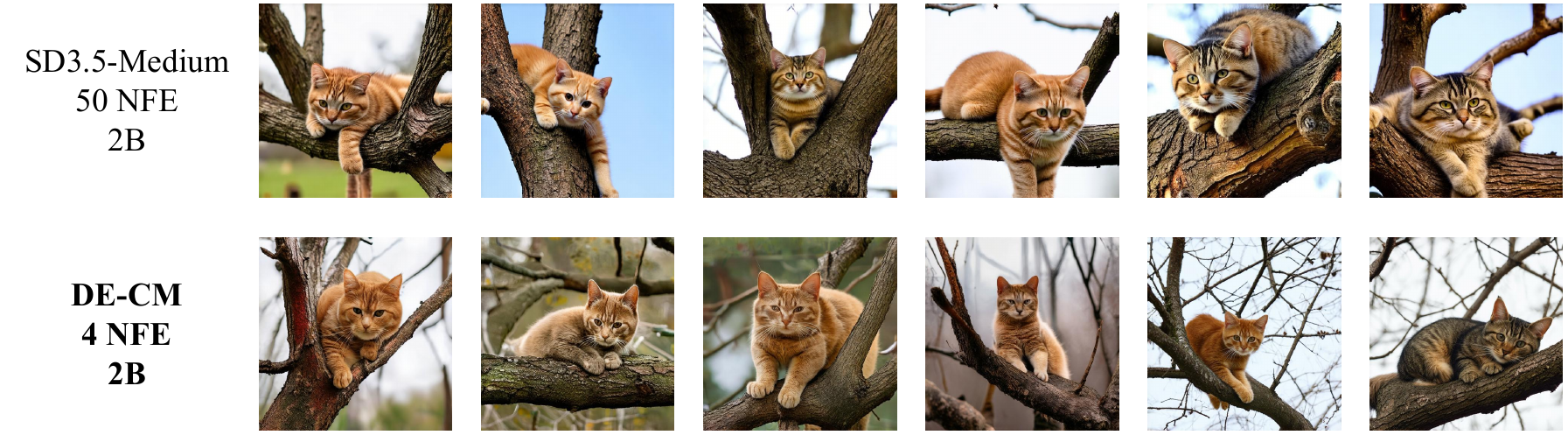}
   \caption{Visual comparison between DE-CM (4 NFE) and its distillation teacher SD3.5-Medium (50 NFE).
   }
   \label{fig:t2i-8}
\end{figure*}

\begin{figure*}[!t]
  \centering
    \includegraphics[width=\linewidth]{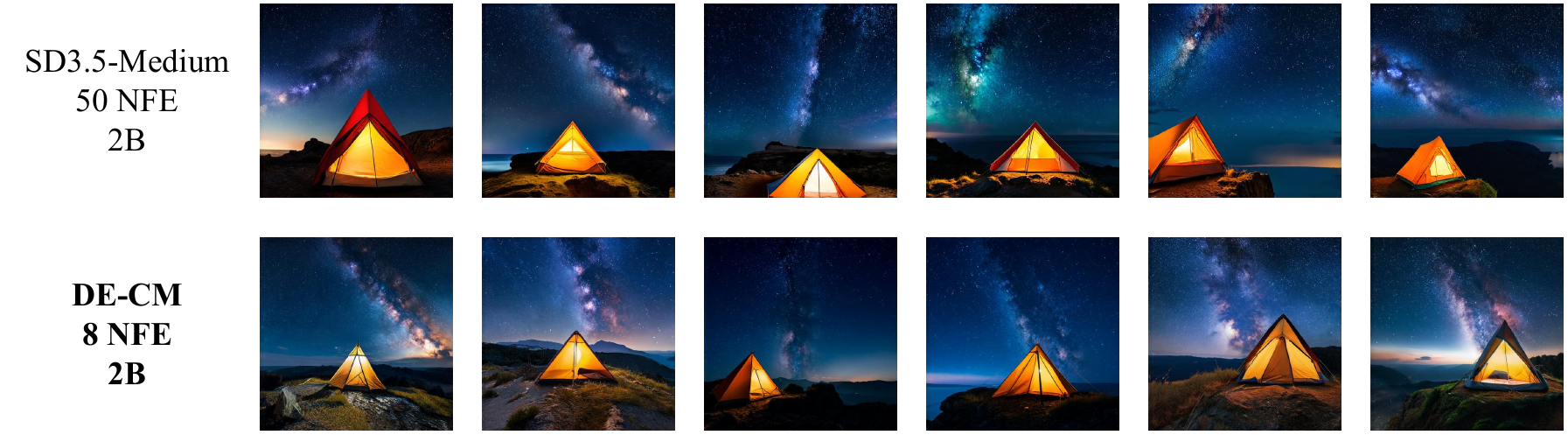}
   \caption{Visual comparison between DE-CM (8 NFE) and its distillation teacher SD3.5-Medium (50 NFE).
   }
   \label{fig:t2i-6}
\end{figure*}

\begin{figure*}[!t]
  \centering
    \includegraphics[width=\linewidth]{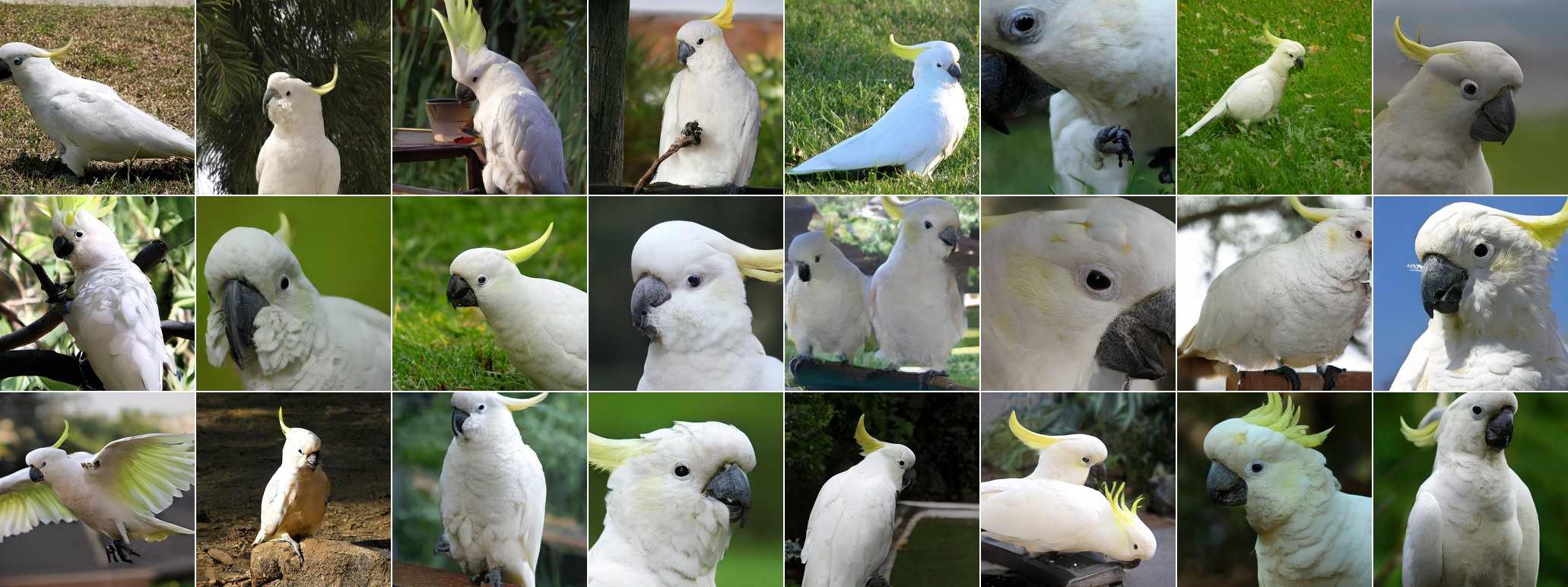}
   \caption{Selected one-step samples generated by our DE-CM, shown for classes
89 (sulphur-crested cockato).
   }
   \label{fig:imagenet-89}
\end{figure*}

\begin{figure*}[!t]
  \centering
    \includegraphics[width=\linewidth]{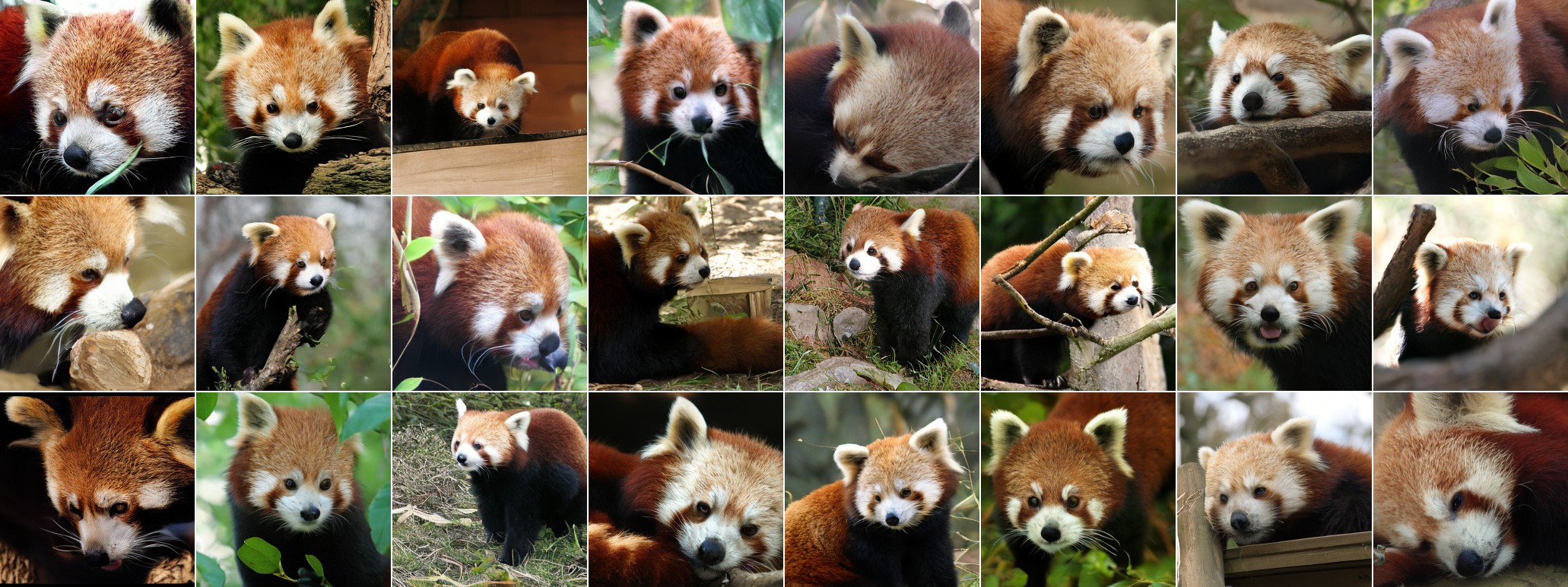}
   \caption{Selected one-step samples generated by our DE-CM, shown for classes
387 (lesser panda).
   }
   \label{fig:imagenet-387}
\end{figure*}

\begin{figure*}[!t]
  \centering
    \includegraphics[width=\linewidth]{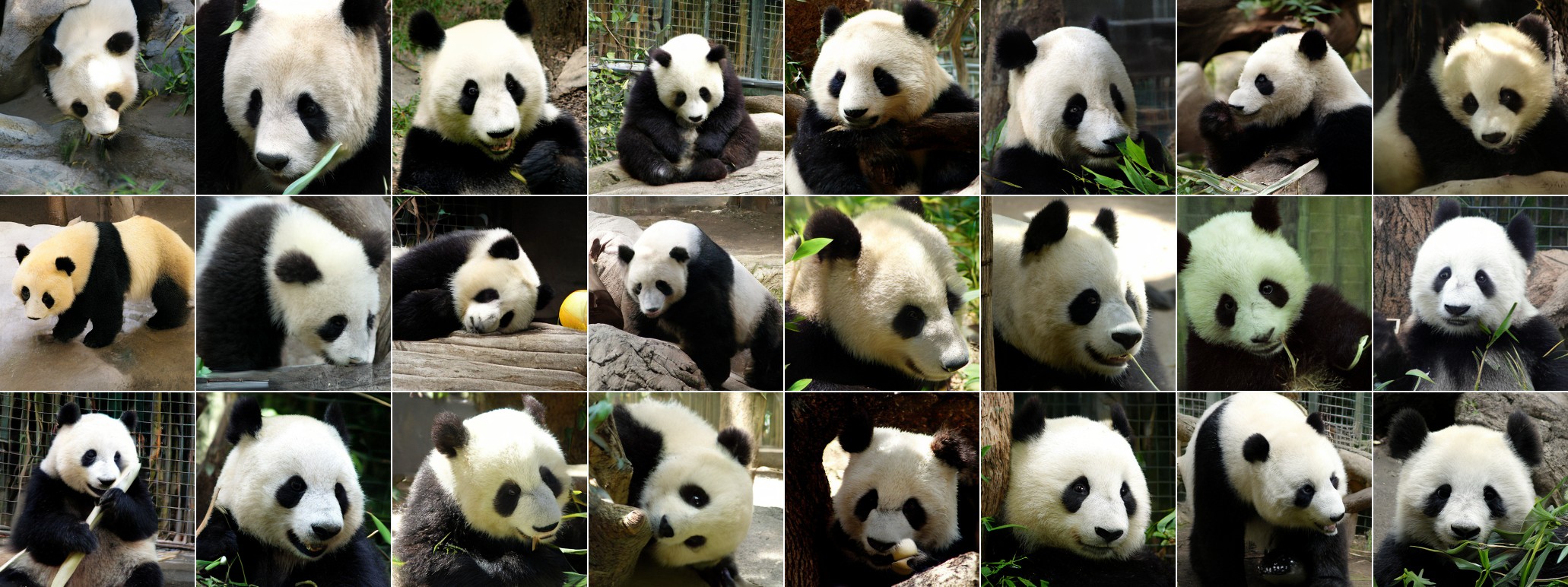}
   \caption{Selected one-step samples generated by our DE-CM, shown for classes
388 (giant panda).
   }
   \label{fig:imagenet-388}
\end{figure*}

\begin{figure*}[!t]
  \centering
    \includegraphics[width=\linewidth]{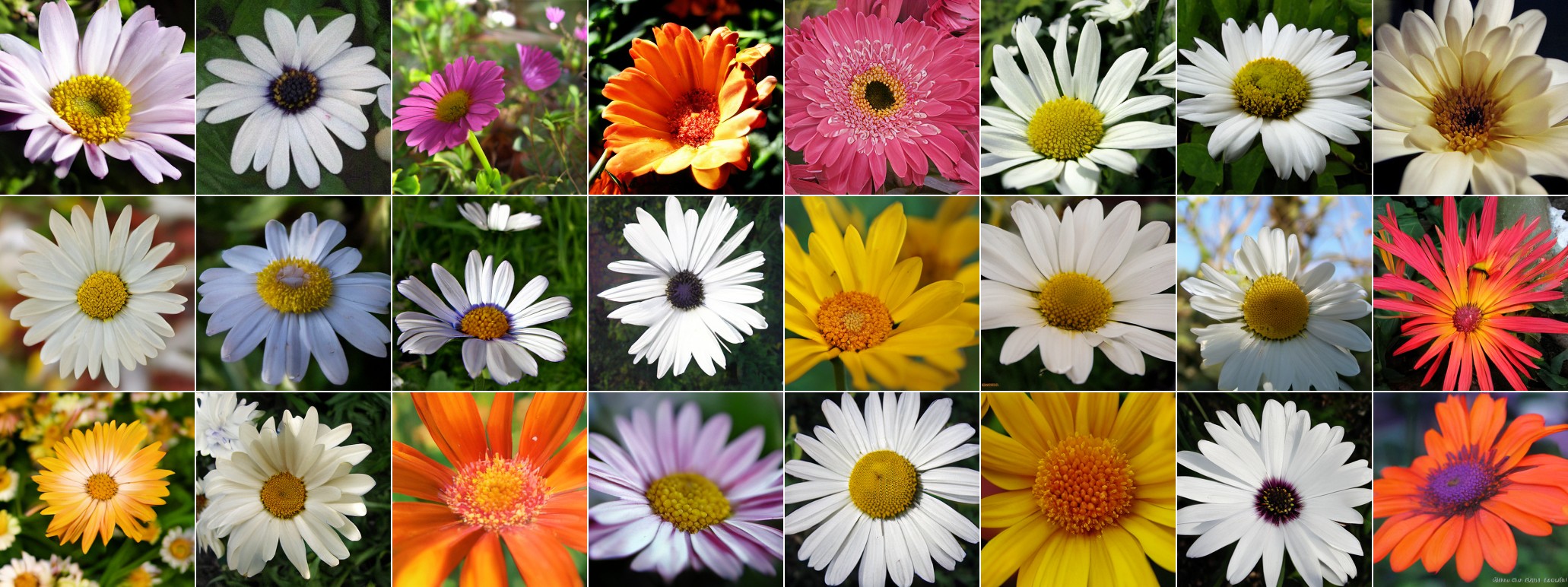}
   \caption{Selected one-step samples generated by our DE-CM, shown for classes
985 (daisy).
   }
   \label{fig:imagenet-985}
\end{figure*}

\begin{figure*}[!t]
  \centering
    \includegraphics[width=\linewidth]{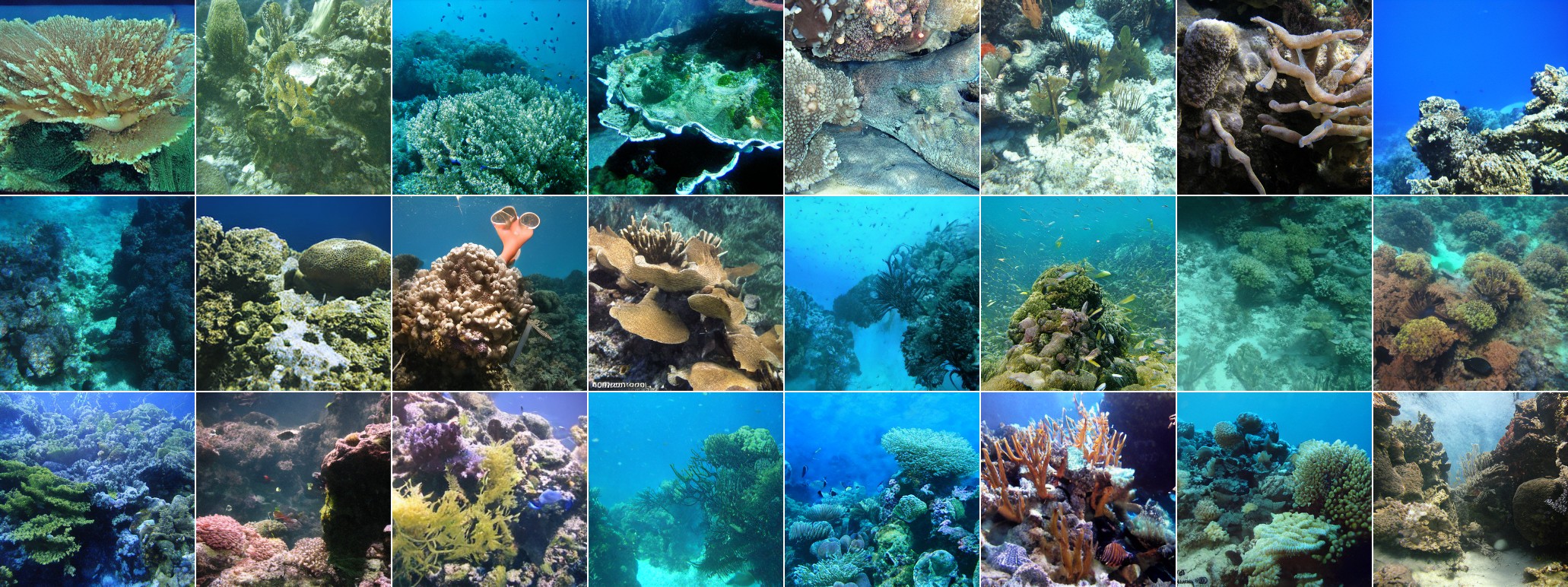}
   \caption{Selected one-step samples generated by our DE-CM, shown for classes
973 (coral reef).
   }
   \label{fig:imagenet-973}
\end{figure*}

\begin{figure*}[!t]
  \centering
    \includegraphics[width=\linewidth]{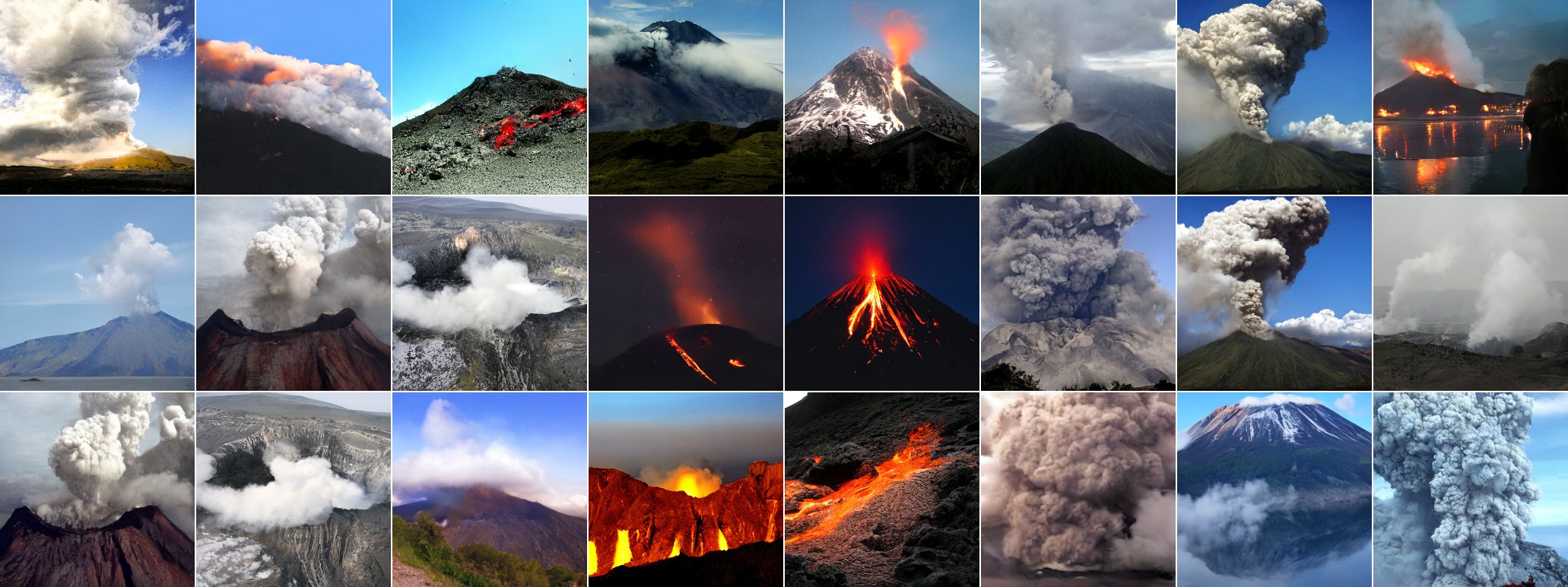}
   \caption{Selected one-step samples generated by our DE-CM, shown for classes
980 (volcano).
   }
   \label{fig:imagenet-980}
\end{figure*}

\begin{figure*}[!t]
  \centering
    \includegraphics[width=\linewidth]{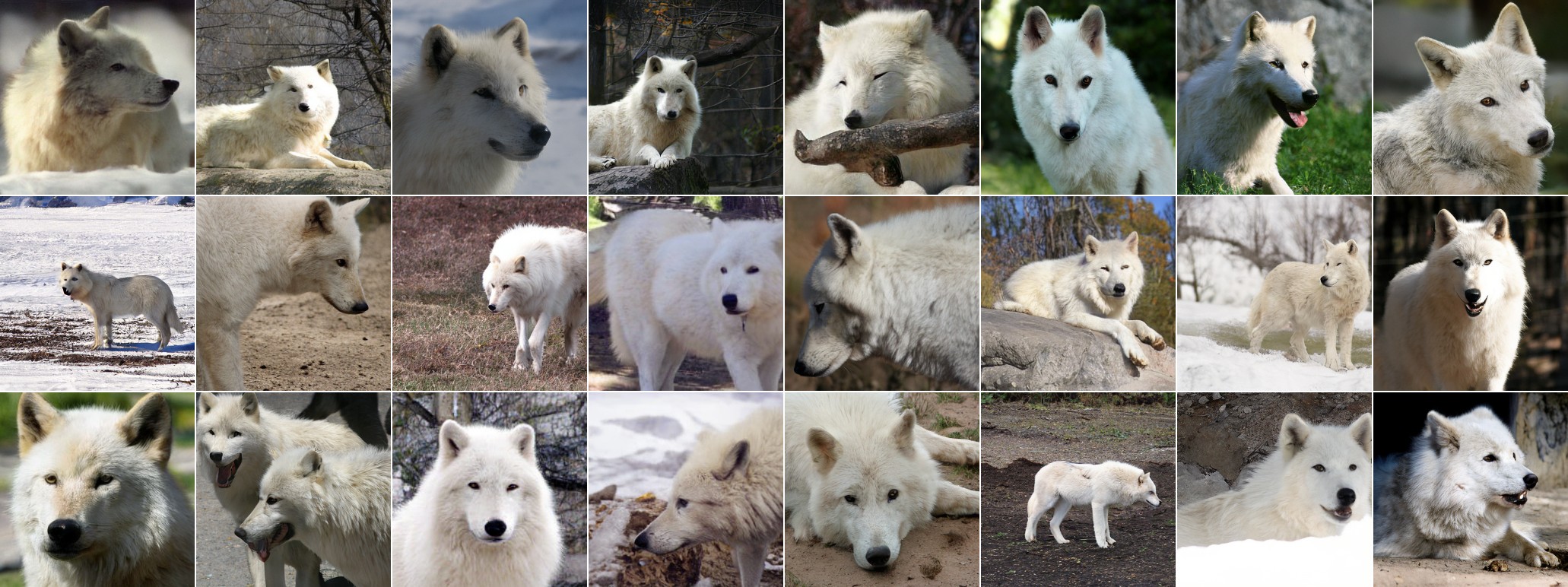}
   \caption{Selected one-step samples generated by our DE-CM, shown for classes
270 (white wolf).
   }
   \label{fig:imagenet-270}
\end{figure*}

\begin{figure*}[!t]
  \centering
    \includegraphics[width=\linewidth]{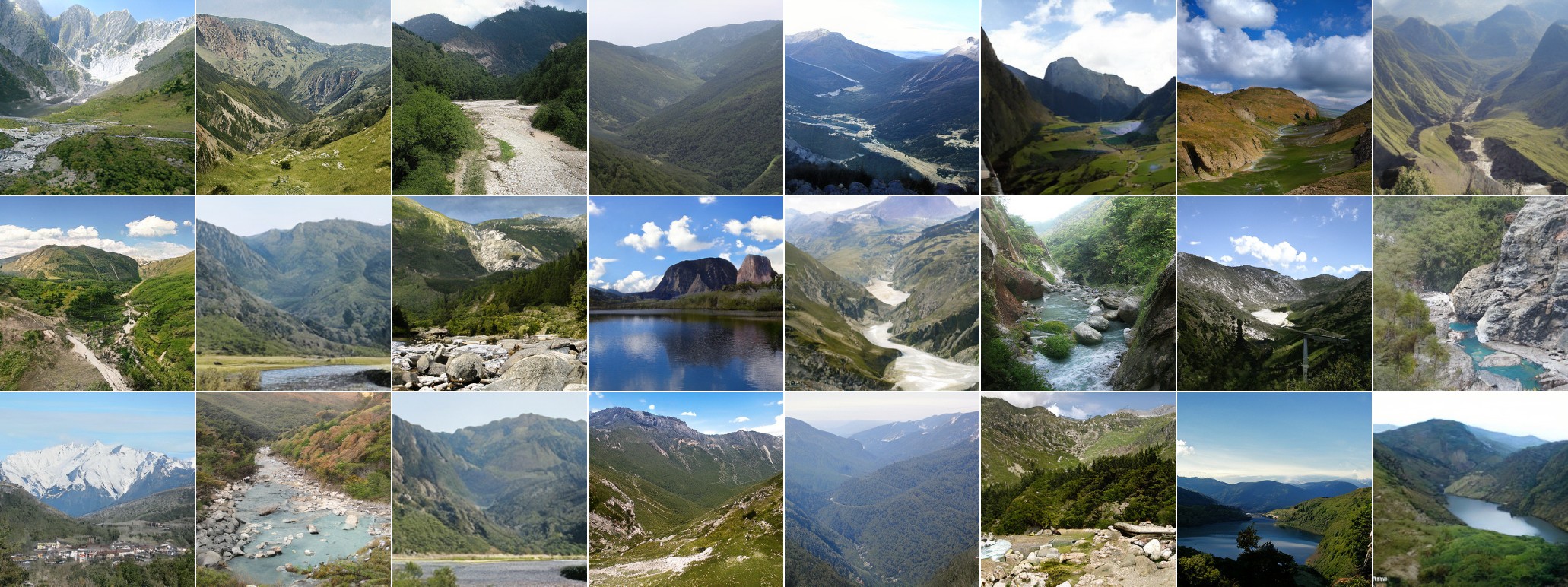}
   \caption{Selected one-step samples generated by our DE-CM, shown for classes
979 (valley).
   }
   \label{fig:imagenet-979}
\end{figure*}

\begin{figure*}[!t]
  \centering
    \includegraphics[width=\linewidth]{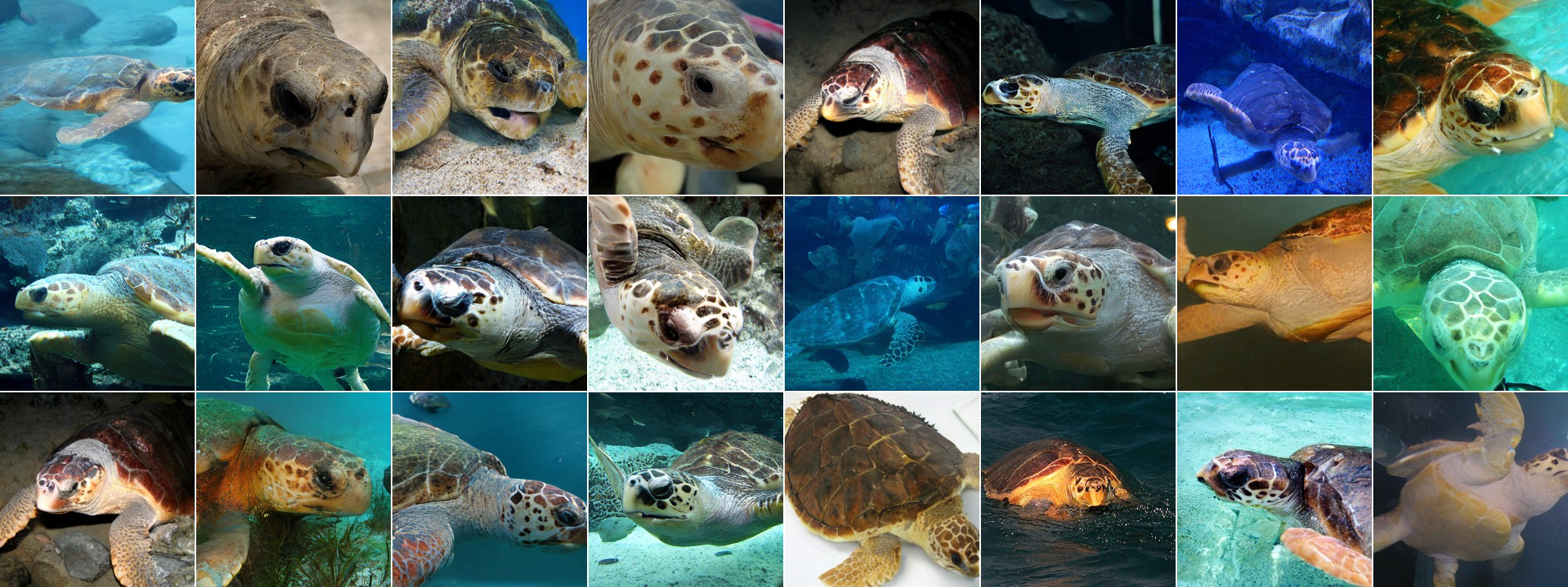}
   \caption{Selected one-step samples generated by our DE-CM, shown for classes
33 (loggerhead).
   }
   \label{fig:imagenet-33}
\end{figure*}

For T2I tasks, we provide additional visual comparisons with existing methods.
\cref{fig:t2i-3} presents a visual comparison of DE-CM with other existing few-step methods under 2 NFE  configuration.
\cref{fig:t2i-4} and \cref{fig:t2i-5} demonstrate a visual quality comparison between our method and large pre-trained models.
\cref{fig:t2i-7}, \cref{fig:t2i-8} and \cref{fig:t2i-6}  show comparisons with the distillation teacher model.
For C2I tasks, in \cref{fig:imagenet-89} - \cref{fig:imagenet-33}, we show additional one-step samples generated by our DE-CM.



\clearpage
\section{Text Prompts Used in Sampling}
\label{ap:text_prompt}
Here we list all the text prompts used to generate the images.

\noindent
In \cref{fig:teaser}:
\begin{itemize}
    \item ``macro photography of a realistic and natural ladybug.''
    \item ``On a still morning, a weathered wooden table holds an empty bowl with cracks, emanating a poignant air of sadness.''
    \item ``A powerful sorceress with arcane tattoos glowing on her skin, casting a spell, surrounded by floating crystals, fantasy concept art, dramatic lighting.''
    \item ``A floating island in the sky with a ancient tree and glowing waterfalls, dreamlike, surreal, soft pastel colors, by James Gurney, fantasy art.''
    \item ``A bustling, miniature village built around giant, colorful mushrooms in a damp forest. Tiny lights glow in the windows. Macro photography, fantasy, highly detailed.''
    \item ``a radiant five-pointed star near a shimmering crescent-shaped moon.''
\end{itemize}
In \cref{fig:t2i}:
\begin{itemize}
    \item ``A mystical forest bathed in moonlight, with glowing blue and green bioluminescent plants. A gentle mist rolls through the trees.''
    \item ``A fluffy ginger cat curled up in a starry night bedroom, in the style of Vincent van Gogh, bold swirling brushstrokes, vibrant blues and yellows.''
    \item ``A stack of pancakes with syrup on top is presented. The pancakes are golden brown and appear to be freshly cooked. The syrup is dripping down the sides of the pancakes.''
    \item ``A cargo ship sails slowly, its hull reflecting a subtle golden hue.''
    \item ``An armored knight lies defeated in a muddy, rainy environment, with an overcast sky and distant landscape.''
\end{itemize}
In \cref{fig:t2i-2}:
\begin{itemize}
    \item ``A close-up shot of a steaming mug of coffee and a knitted wool sweater on a rustic wooden table, autumn leaves scattered around. Soft focus, texture emphasis, cozy.''
    \item ``A bioluminescent forest, glowing jellyfish, floating gently in a misty twilight. Fantasy concept art.''
    \item ``An ancient spellbook with a glowing crystal floating, in a dark, dusty library, magical aura, intricate details.''
    \item ``fantasy, a majestic sky filled with stars and galaxies, over looking a serene lake.''
    \item ``Macro detail of a dragon's scale, each one like an intricate piece of jeweled armor, reflecting fiery light. Hyperdetailed, Unreal Engine 5.''
\end{itemize}
In \cref{fig:t2i-3}:
\begin{itemize}
    \item ``A brown leather satchel with silver buckles is displayed on a black surface with a white background.''
    \item ``A beautiful girl with long silver hair and blue eyes, wearing a elegant white dress, in a field of glowing flowers, anime style, vibrant colors, detailed background.''
    \item ``A floating island in the sky with a ancient tree and glowing waterfalls, dreamlike, surreal, soft pastel colors, by James Gurney, fantasy art.''
    \item ``macro photography of a realistic and natural ladybug.''
\end{itemize}
In \cref{fig:t2i-4} and \cref{fig:t2i-5}:
\begin{itemize}
    \item ``A powerful sorceress with arcane tattoos glowing on her skin, casting a spell, surrounded by floating crystals, fantasy concept art, dramatic lighting.''
    \item ``A bustling, miniature village built around giant, colorful mushrooms in a damp forest. Tiny lights glow in the windows. Macro photography, fantasy, highly detailed.''
\end{itemize}
In \cref{fig:t2i-7} , \cref{fig:t2i-8} and \cref{fig:t2i-6}:
\begin{itemize}
    \item ``The scene shows a wide panoramic view of a marsh at sunset. The golden light from the sun casts a warm glow over the water.''
    \item ``A cat draped over a barren tree branch.''
    \item ``A lone, cozy A-frame tent pitched on a cliff edge under a breathtaking starry night and the Milky Way. The tent glows with warm light from within. Astrophotography, serene, majestic.''
\end{itemize}

\end{document}